\documentclass[10pt,twocolumn,letterpaper]{article}
\pdfoutput=1
\RequirePackage[format=plain,labelformat=simple,labelsep=period,font=small,compatibility=false]{caption}
\RequirePackage[font=footnotesize,skip=3pt,subrefformat=parens]{subcaption}
\usepackage{lmodern}
\usepackage[T1]{fontenc}
\usepackage[utf8]{inputenc}
\usepackage[ngerman,english]{babel}
\usepackage{graphicx}
\graphicspath{{./}{./figs/}}
\usepackage[usenames]{xcolor}
\definecolor{fgreen}{rgb}{0.1,0.5,0.2}
\definecolor{grape}{rgb}{0.42,0.2,0.38}
\definecolor{marine}{rgb}{0,0.2,0.6}
\usepackage{enumitem}
\usepackage{listings}
\lstnewenvironment{lstcode}[1][]{\lstset{language=Python,#1}}{}
\newcommand{\lsti}{\lstinline}
\PassOptionsToPackage{hyphens}{url}
\usepackage[bookmarks,bookmarksnumbered,pagebackref,breaklinks,colorlinks=true,linkcolor=grape,citecolor=fgreen,urlcolor=marine,pdfa=true]{hyperref}
\usepackage{url}
\usepackage{verbatim}
\usepackage[font=small]{caption}
\usepackage[capitalize]{cleveref}

\newcommand{\myTitle}{Deep Learning Reproducibility and Explainable AI (XAI)}
\hypersetup{
  pdfauthor={Dr. Anastasia-Maria Leventi-Peetz <leventi@bsi.bund.de>},
  pdftitle={\myTitle},
  pdfkeywords={artifical intelligence, machine learning, deep learning, artificial neural networks, convolutional neural networks, reproducibility, robustness, explainability, determinism},
  pdflang={en-US}
}
\begin{document}
\title{\bf \myTitle}
\author{A.-M. Leventi-Peetz\\
Federal Office for Information Security (BSI) Germany\\
{\tt\small leventi@bsi.bund.de}
\and
T. Östreich\\
Federal Office for Information Security (BSI) Germany
}
\date{}
\maketitle

\begin{abstract}
The nondeterminism of Deep Learning (DL) training algorithms and its influence on the explainability of neural network (NN) models are investigated in this work with the help of image classification examples. To discuss the issue, two convolutional neural networks (CNN) have been trained and their results compared. The comparison serves the exploration of the feasibility of creating deterministic, robust DL models and deterministic explainable artificial intelligence (XAI) in practice. Successes and limitation of all here carried out efforts are described in detail. The source code of the attained deterministic models has been listed in this work. Reproducibility is indexed as a development-phase-component of the Model Governance Framework, proposed by the EU within their excellence in AI approach. Furthermore, reproducibility is a requirement for establishing causality for the interpretation of model results and building of trust towards the overwhelming expansion of AI systems applications. Problems that have to be solved on the way to reproducibility and ways to deal with some of them, are examined in this work.
\end{abstract}

\section{Introduction}
\label{sec:intro}
\subsection{Reproducible ML models}
The reproducibility of ML models is a subject of debate
with many aspects under investigation by researchers and practitioners
in the field of AI algorithms and their applications.
Reproducibility refers to the ability to
duplicate prior results using the same means as used in the
original work, for example the same program code and raw data. However, ML experiences what is called a reproducibility
crisis and it is difficult to reproduce important ML results, some also described as \emph{key results}~\cite{reprocrisisA, reprocrisis, keyrepro,MallReporCrisis2019}.
Experience reports refer to many publications as being not replicable, or being statistically insignificant, or suffering from narrative fallacy~\cite{dennybritzai-replication-incentives}.
Especially Deep Reinforcement Learning has received a lot of attention with many papers~\cite{dennybritzai-replication-incentives,Khetarpal2018Reproducibility,Islam2017Reproducibility,Engstrom2020ImplementationMatters} and blog posts~\cite{DRdoesnotworkyet} investigating the high variance of some results.
Because it is difficult to decide which ML results are trustworthy and generalize to real-world problems, the importance of reproducibility is growing.
A common problem concerning reproducibility is when the code is not
open-sourced. The review of 400 publications of two top AI conferences in the last years, showed that only 6\,\%
of them shared the used code, one third shared the data on which algorithms
were tested and half shared pseudocode~\cite{Gundersen400, missingdataHutson}.
Initiatives like the \emph{2019 ICLR reproducibility challenge}~\cite{ICLRreproC2019} and
the \emph{Reproducibility Challenge of NeurIPS 2019}~\cite{NeurIPS2019RC,ipsReproducibility19},
that invite members of the AI community to reproduce papers accepted at
the conference and report on their findings via the OpenReview
platform ({\small\url{https://openreview.net/group?id=NeurIPS.cc/2019/Reproducibility_Challenge}}), demonstrate an increasing intention to make machine learning trustworthy
by making it computationally reproducible~\cite{HeilNature}.
Reproducibility is important for many reasons:
For instance, to quantify progress in ML, it has to be certain that noted
model improvements originate from true innovation and are not the  sheer product of uncontrolled randomness~\cite{dennybritzai-replication-incentives}.  
Also from the
development point of view, adaptations of models to changing
requirements and platforms are hardly possible in the absence of
baseline or reference code, which works according to agreed upon expectations.
The latter could get transparently extended or changed before tested to meet
new demands.
For ML models, it is the so named \emph{inferential reproducibility} which is important as a requirement and states
that when the inference procedure is repeated, the results should be
qualitatively similar to those of the original procedure~\cite{keyrepro}.
However, training reproducibility is also
a necessary step towards the formation of a systematic framework for an
end-to-end comparison of the quality of ML models. To our knowledge such a framework does not yet exist and it should be
essential if criteria and guarantees regarding the quality of ML models
have to be provided.
Security and safety considerations are inevitably involved: For instance, when a model executes a
pure classification exercise, deciding for example if a test image shows a cat
or a dog, it is not necessarily critical when the model’s decision turns out to be
wrong. If however the model is incorporated into a clinical decision-making
system, that helps make predictions about pathologic conditions on the basis of patients’ data, or is part of an automated driving system (ADS) which
actively decides if a vehicle has to immediately stop or keep speeding, then
the decision has to be verifiably correct and understandable at every stage of its formation.
The increasing dependency on ML for decision making leads to an
increasing concern that the integration of models which have not been
fully understood can lead to unintended consequences~\cite{repro-towards-data-science}.
\subsection{Factors hindering training reproducibility}
It is well known that when a model is trained
again with the same data it can produce different predictions~\cite{differentResultsJason, differentResultsJason1}.
To the reasons that make reproducibility
difficult there belong: different problem formulations, missing compatibility between DNN-architectures, missing appropriate benchmarks, different OS,
different numerical libraries, system architectures
or software environments like the
Python version etc. 
Reproducibility as a basis 
for the generation of sound  explanations and interpretations of model decisions
is also essential in view of the immense computational effort and costs
involved when applying or adapting algorithms, often without specific
knowledge about the hardware, the parameter-tuning and the energy
consumption demanded for the training of a model,
which at the end might lead to inconclusive results.
Furthermore, it is also difficult to train models
to expected accuracy even when the program code and the training data are available.
Changes in TensorFlow, in GPU drivers, or even slight changes in the datasets, can hurt accuracy in subtle ways~\cite{PWardenReproCrisis,determinedAI}. In addition, many ML models are trained on restricted datasets, for example those containing sensitive patient information, that can’t be made publicly available~\cite{Crisisin3EasySteps}.
When privacy barriers are important considerations for data sharing, so called replication processes have to be used, to investigate the extent to which the
original model generalizes to new contexts and new data populations, and
decide whether similar conclusions to those of the original model can be delivered.
However, there exist also certain unique challenges which ML reproducibility
poses.
The training of ML models makes use of randomness, especially for DL, usually employing stochastic gradient descent, regularization techniques etc.~\cite{ChallengestoReproducibility}.
Randomized procedures result in different final values for the model parameters every time the code is executed. One can set all possible random seeds, however additional parameters, commonly named silent parameters, associated with modern deep learning, have been found to also have a profound influence on both model performance and reproducibility.
High-level frameworks like Keras are reported to hide low-level implementation details and come with implicit hyperparameter choices already made for the user. Also hidden bugs in the source code can lead to different outcomes in dependence of linked libraries and different execution environments.
Moreover, the cost to reproduce state-of-the-art deep learning models is often extremely high. In natural language processing (NLP), \emph{transformers} require huge 
amounts of data and computational power and can have in excess of 100 billion trainable parameters.
Large organizations produce models (like \mbox{OpenAI}'s GPT-3) which can cost millions of dollars in computing power to train~\cite{Crisisin3EasySteps,ChallengestoReproducibility,untoldstoryGPT}.
To find the transformer that achieves the best predictive performance for a given application, \emph{meta-learners} test
thousands of possible configurations. The cost to reproduce one of the
many possible transformer models has been estimated to range from 1~million to 3.2~million USD with usage of publicly available cloud computing resources~\cite{emissions,ChallengestoReproducibility}. This process is estimated to generate CO$_2$ emissions with a volume which amounts to the fivefold of emissions of an average car, generated over its entire lifetime on the road. The environmental implications attached to reproducibility endeavors of this range are definitely prohibitive~\cite{ChallengestoReproducibility}.
As possible solution to this
problem, there has been proposed the option to let expensive large models get produced only once, while adaptations of these models for special applications should be made transparent and
reproducible with the use of more modest resources~\cite{ChallengestoReproducibility}.
\subsection{Organization and aim of this work}
The majority of methods for explainable AI are attribute based, they
highlight those data features (attributes), that mostly contributed to the
model's prediction or decision.
Convolutional neural networks (CNN, or ConvNet)  are state-of-the-art architectures, for which visual explanations can be
produced, for example with the Gradient-weighted Class Activation Mapping method (Grad-CAM)~\cite{GradCAMselvaraju,CholletGradCAMcav2021}, which is also the method used in this work.
In the second part of this work, Grad-CAM explanations for two pre-trained
and established CNN models, which use TensorFlow, will be discussed
with focus on the differences of their results, when the same test-data are given as input.
It is well known that when different explainability methods are applied on a neural network, different results are to be expected. The fact that a single
explainability method, when applied on two similar CNN-architectures, can produce different results for the same test-data, has received less attention in the literature but is worth to analyze in the reproducibility context.
In the third part, the own implementation, training and results of two relatively
simple CNN models are discussed.
Differences of the Grad-CAM-explanations for identical images
classified with these two networks are analyzed, with special focus on the influence
of the computing infrastructure on the model execution.
The efforts to render these two models deterministic
are described in \cref{sec:DetSelfTrainMod} in detail, again with special focus on the influence of the computing infrastructure on the results.
Success and limitations are noted, the partly achieved  deterministic code is listed.
It is worth mentioning that different
behaviors across versions of TensorFlow, as well as across different
computational frameworks are documented to be normally
expected. TensorFlow warns that floating point values computed by ops,
may change at any time and users should rely only on \emph{approximate
accuracy and numerical stability, not on the specific bits
computed}. There could be found no experience reports, as to how a change of
\emph{specific bits} could influence ML results, for
instance in worst case by altering the network's classification or its
explanation, or both.  According to TensorFlow, changes to numerical formulas
in minor and patch releases, should result in comparable or improved
accuracy of specific formulas, with the \emph{caution} that this might
\emph{decrease the accuracy} for the overall system. Also models implemented in one version of TensorFlow, cannot run with next subversions and versions of TensorFlow. Therefore published code which was once proved to work, is possibly not to use again within
short time after its creation. To run more
than one subversions on the same system, when using graphic HW support,
was not possible. This work aims at drawing attention to the challenges that adhere to
creating reproducible training processes in Deep Learning and demonstrates practical steps towards reproducibility, discussing their present limitations.
In \cref{sec:Concl} conclusions of this work and views towards future investigations in the same direction are presented in a summary. It has to be noted that the impact of what is called \emph{underspecification}, whereby
the same training processes produces multiple machine-learning models which demonstrate
differences in their performance, is out of scope of this work~\cite{underspecification}.
%
%
\section{Grad-CAM NN-Explanations}
\subsection{Network architectures and HW}
\label{s:archHW}
Convolutional neural networks, originally developed
for the analysis and classification of objects in digital images, represent the core of most state-of-the-art computer vision solutions for a wide variety of tasks~\cite{rethinkingSzegedy}. A brief but comprehensive history of CNN can be found in many
sources, for example in~\cite{CholletXCeption-2017}, whereby the tendency has always been towards making CNN increasingly deeper. Developments of the last years have led to the \emph{Inception
architecture}, which incorporates the so called \emph{Inception
modules}, that exist already in several different
versions. A new architecture, which 
instead of stacks of simple convolutional networks, contains stacks of convolutions itself,
was proposed by François Chollet with his \emph{Extreme Inception} or \emph{Xception} model. Xception was proved to be capable of learning richer representations with less parameters~\cite{CholletXCeption-2017}.
Chollet delivered the Xception improvements to the Inception family of
NN-architectures, by entirely replacing Inception modules with depthwise separable
convolutions. Xception also uses residual connections, placed in all flows of the network~\cite{CholletXCeption-2017,He2016DeepResLearn}.
The role of residuals was observed as especially important for the convergence
of the network~\cite{xception-datatoward}, however Chollet moderates this importance, because non-residual
models have been benchmarked with the same optimization configuration as the residual ones, which leaves the possibility open, that another configuration might have proved the non-residual version better~\cite{CholletXCeption-2017}.
Finally, the building of the improved Xception models was made possible because an
efficient depthwise convolution implementation became available in
TensorFlow.
The Xception architecture has a
similar number of parameters as Inception V3. Its performance 
however has been found to be better than that of Inception, according to tests
on two large-scale image classification tasks~\cite{CholletXCeption-2017}.
For practical tests in this work, Inception V3 and Xception
have been chosen for results comparisons.
The two networks are pretrained on a trimmed list of the ImageNet dataset,
so as to be able to recognize one thousand non-overlapping object classes~\cite{CholletXCeption-2017}.
\begin{description}[itemindent=-2mm]
\item[Inception V3]
  The exact description of the network, its parameters and performance are given in the work of Christian Szegedy~\cite{szegedyInceptionV4}. The description of the training infrastructure refers to a system of 50 replicas, (probably identical systems), running each on a NVidia Kepler GPU, with batch size 32, for 100 epochs. The time duration of each epoch is not given.
\item[Xception]
  Chollet has used 60 NVIDIA K80 GPUs for the training, which took a duration
  of 3 days time. The number of epochs is not given. The network and technical details about the training are listed in the original work~\cite{CholletXCeption-2017}.
\end{description}
%
Xception has a similar number
of parameters (ca.~23 million) as Inception V3 (ca.~24 million).
The HW execution environments employed for the here described experiments are the following:
\begin{itemize}\setlength{\itemsep}{-0.4ex}\small
  \item
  HW-1:
  GPU: NVIDIA TITAN RTX: 24 GB (GDDR6), 576 NVIDIA Turing mixed-precision Tensor Cores, 4608 CUDA Cores.
\item
  HW-2:
  CPU: AMD EPYC 7502P 32-Core, SMT, 2 GHz (T: 2.55 GHz), RAM 128 GB.
\item
  HW-3:
  GPU: NVIDIA GeForce RTX 2060: 6 GB (GDDR6), 240 NVIDIA Turing mixed-precision Tensor Cores, 1920 CUDA Cores.
 \item
  HW-4:
  CPU: AMD Ryzen Threadripper 3970X 32-Core, SMT, 3.7 GHz (T: 4.5 GHz), RAM 256 GB.
  \item
  HW-5:
  CPU: AMD Ryzen 7 5800X 8-Core, SMT, 3.8 GHz (T: 4.7 GHz), RAM 64 GB.
\end{itemize}
Each of the pretrained models is verified to deliver the same results
for all here considered CPU or GPU different execution
environments. The classifications and the according
network explanations are deterministic when performed under laboratory conditions,
as also expected. Plausibility and stability issues of the explanations will be mentioned parallel to the tests. 
\subsection{Inception V3}
\label{s:inceptionv3exam}
\begin{figure}[htbp]\setlength{\tabcolsep}{0.7mm}
  \centering
  \begin{tabular}{cc}
    \includegraphics[width=0.48\linewidth, viewport=70 10 370 240, clip]{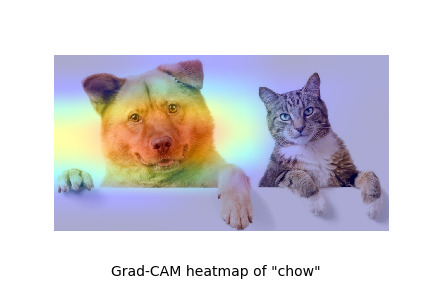} &
    \includegraphics[width=0.48\linewidth, viewport=70 10 370 240, clip]{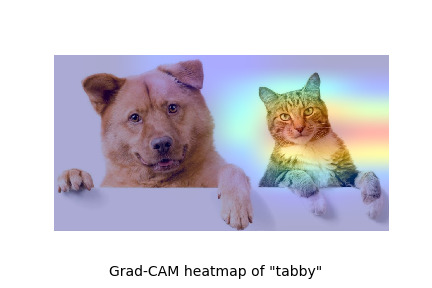} \\
    {\small (a)} & {\small (b)} \\
  \end{tabular}
  \caption{\label{fig:inc-chowtabby}\emph{chow-cat}: Grad-CAM explanations of Inception V3 for the identification of the dog ``chow'' (a), in the first place on the top-predictions-list and the cat ``tabby''(b), in the third place on the top-predictions-list. The second place occupies a ``Labrador dog''.}
\end{figure}

\begin{figure}[htbp]\setlength{\tabcolsep}{0.7mm}
  \centering
  \begin{tabular}{ccc}
    \includegraphics[width=0.32\linewidth, viewport=70 10 370 240, clip]{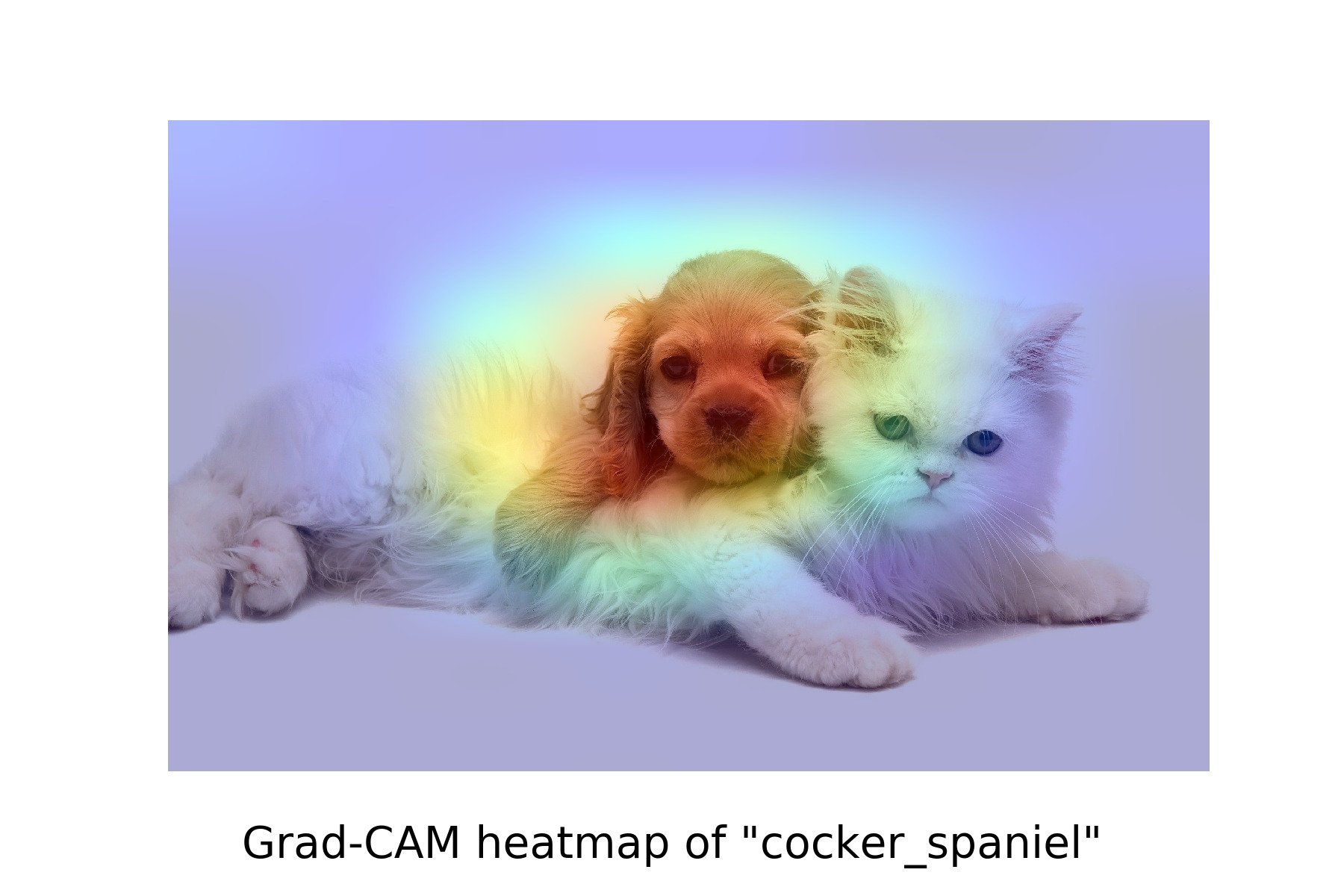} &
    \includegraphics[width=0.32\linewidth, viewport=70 10 370 240, clip]{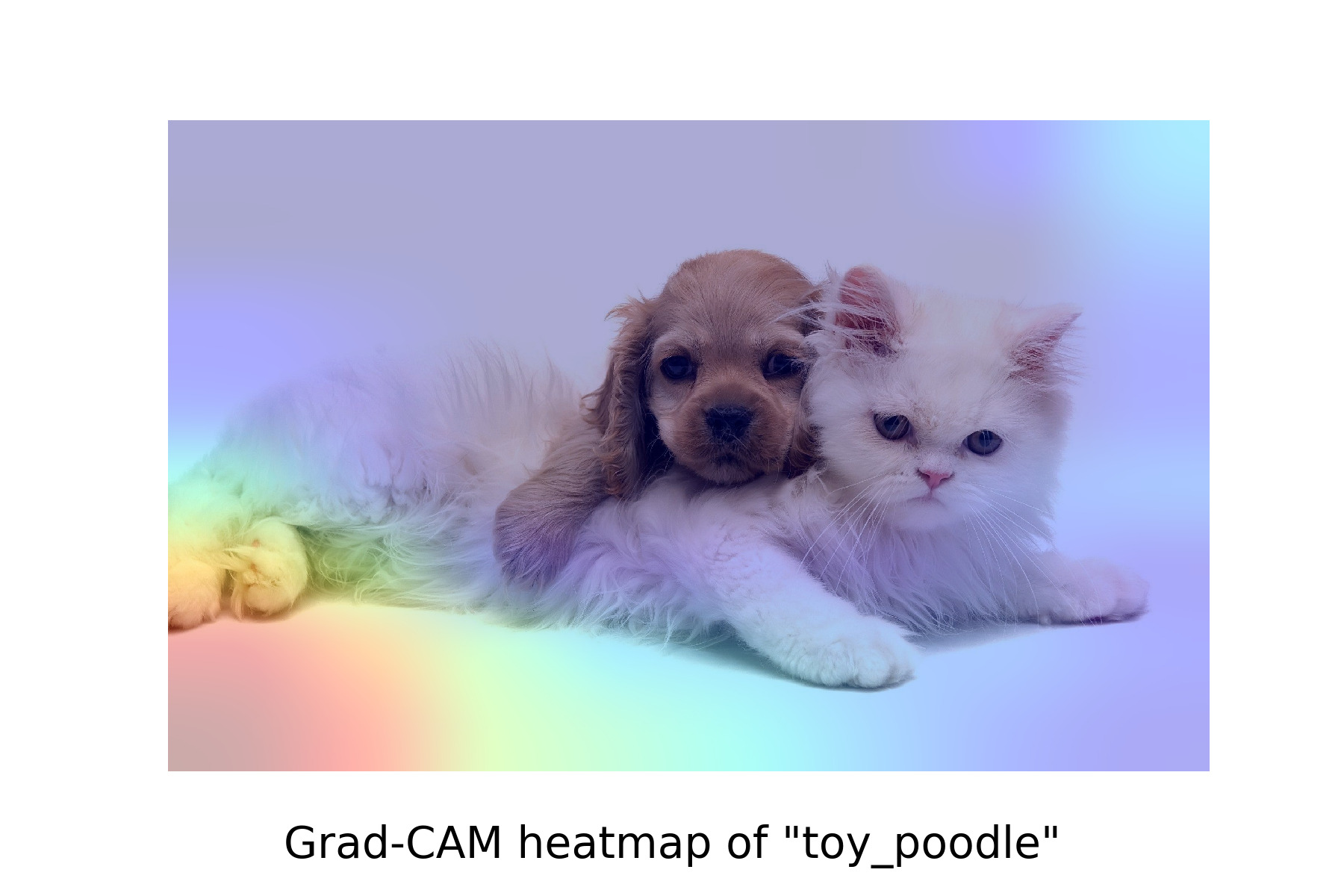} &
    \includegraphics[width=0.32\linewidth, viewport=70 10 370 240, clip]{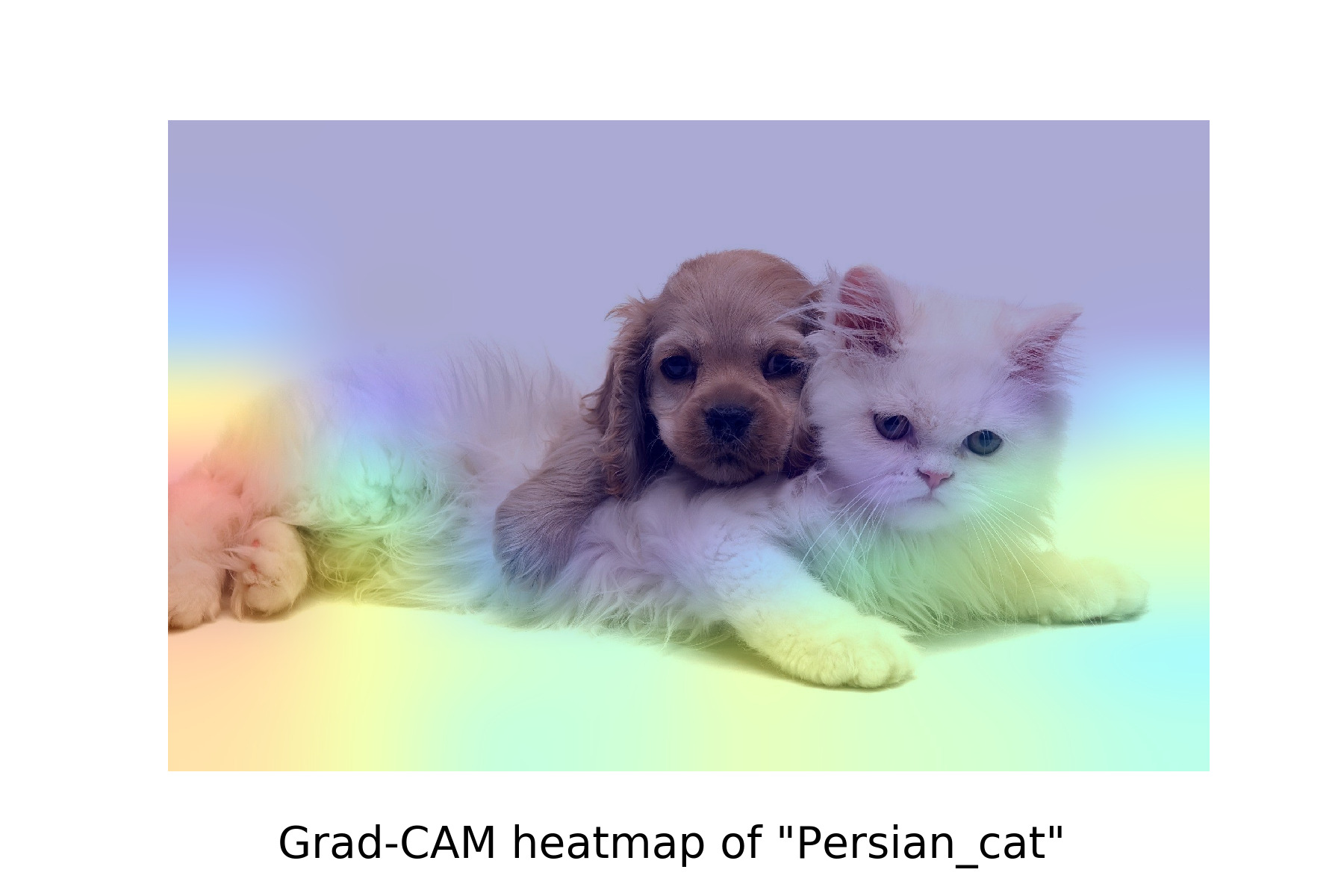} \\
    {\small (a)} & {\small (b)} & {\small (c)} \\
  \end{tabular}
  \caption{\label{fig:inc-spanielPcat}\emph{spaniel-kitty}: Grad-CAM explanation of Inception V3 for the identification of the ``cocker spaniel'' (a), the ``toy poodle'' (b) and the ``Persian cat'' (c), see Table \ref{tb:inseption-scores}.}
\end{figure}

In this part examples of predictions, calculated with the Inception V3 network are discussed.
In Fig.~\ref{fig:inc-chowtabby} (a) and (b) respectively, there are depicted activation heatmaps which have been produced to identify those regions of the
image \emph{chow-cat}, that correspond to the dog (``chow'') and the cat (``tabby'') respectively. Identical respective accuracies have been calculated for each classification independent of the employed HW, as was verified by the tests performed with all HW-environments listed at the end of \ref{s:archHW}.  The ``chow'' has been predicted with 30\,\%
probability and stands in the first place on the top-predictions-list, while the cat gets
the third position with a probability of 2.4\,\%.
%
In Fig.~\ref{fig:inc-spanielPcat}, heatmaps produced by the
identification of the ``cocker spaniel'' (a), the ``toy poodle'' (b), and the ``Persian cat'' (c) respectively, have been demonstrated for the image \emph{spaniel-kitty}.
%
%
%
\begin{table}[htbp]
  \caption{\label{tb:inseption-scores} Inception V3: Classification Probabilities for the image \emph{spaniel-kitty}, see Fig.~\ref{fig:inc-spanielPcat}.}
  \begin{minipage}[t]{\linewidth}\small
  \centering
  \vspace{2mm}
  \begin{tabular}{| l |c | c | c |}
  \hline
    &  Class & HW-2 & HW-1 \\
  \hline
  1 & cocker spaniel & 0.56762594               &  0.56761914\\
  2 & toy poodle &  0.08013367            &  0.08014054 \\
  3 & clumber  &  0.02106595   & 0.02107035 \\
  4 & Dandie Dinmont & 0.01964365 & 0.01964012 \\
  5 & Pekinese       & 0.01867950 & 0.01868443 \\
  6 & miniature poodle & 0.01846011 & 0.01846663 \\
  7 & Blenheim spaniel & 0.01425239 & 0.01424699 \\
  8 & Maltese dog & 0.01124849 & 0.01124578 \\
  9 & Chihuahua & 0.01103328 & 0.01103479 \\
  10 & Norwich terrier & 0.00741338 & 0.00741514 \\
  11 & Sussex spaniel & 0.00703137 & 0.00703068 \\
  12 & Yorkshire terrier & 0.00689254 & 0.00689154 \\
  13 & Norfolk terrier & 0.00662250 & 0.00662296 \\
  14 & Lhasa & 0.00609926  & 0.00609862 \\
  15 & Pomeranian & 0.00608485 & 0.00608792  \\
  16 & Persian cat & 0.00489533 & 0.00489470 \\
  17 & golden retriever & 0.00428663 & 0.00428840\\
  \hline
  \end{tabular}
  \end{minipage}%
\end{table}
%
%
\begin{table}[htbp]
  \caption{\label{tb:inseption-scores1}Inception V3: Classification Probabilities for the image \emph{spaniel-kitty-paws-cut} .}
  \begin{minipage}[t]{\linewidth}\small
  \centering
  \vspace{2mm}
  \begin{tabular}{| l |c | c | c|}
  \hline
                 &  Class & HW-2 & HW-1 \\
  \hline
 1 & cocker spaniel & 0.43387938          &  0.43393657 \\
 2 & Persian cat &  0.03001592           &   0.03000891 \\
 3 & Pekinese  &  0.02654952             &   0.02654130\\
 4 &  toy poodle & 0.01810920             &   0.01810851 \\
 5 &  Dandie Dinmont & 0.01457902             &   0.01457707 \\
 6 & Sussex spaniel & 0.01415453            &   0.01415372 \\
 7 & Golden retriever &  0.01363987          &   0.01363916 \\
 8 & Miniature poodle &  0.01088122            &   0.01088199 \\
  \hline
  \end{tabular}
  \end{minipage}%
\end{table}
In Table~\ref{tb:inseption-scores} there are
listed the scores of the first 17 classes on the top-predictions-list,
as calculated in two HW executions (HW-1,HW-2).
The prediction scores are almost identical, as is obvious by comparing the columns in Table~\ref{tb:inseption-scores}, while in the few cases, when slight differences exist
in the probability values, these
differences appear only after the fourth decimal place. The ``cocker spaniel'' is the top prediction and represents actually the correct classification of the dog race, predicted with a probability of almost 57\,\%, while the ``Persian cat'' in place 16 of the list, which is also a correct prediction, has a probability of approximately 0.5\,\%.
The ``toy poodle'' with 8.0\,\% probability stands in the second place on the list,
while the rest of list places, down to place sixteen of the ``Persian cat'', are all occupied by dog races (see
Table~\ref{tb:inseption-scores}).

\subsubsection{Soundness and stability of explanations}
\begin{figure}[htbp]\setlength{\tabcolsep}{0.7mm}
  \centering
  \begin{tabular}{ccc}
    \includegraphics[width=0.32\linewidth, viewport=70 10 370 240, clip]{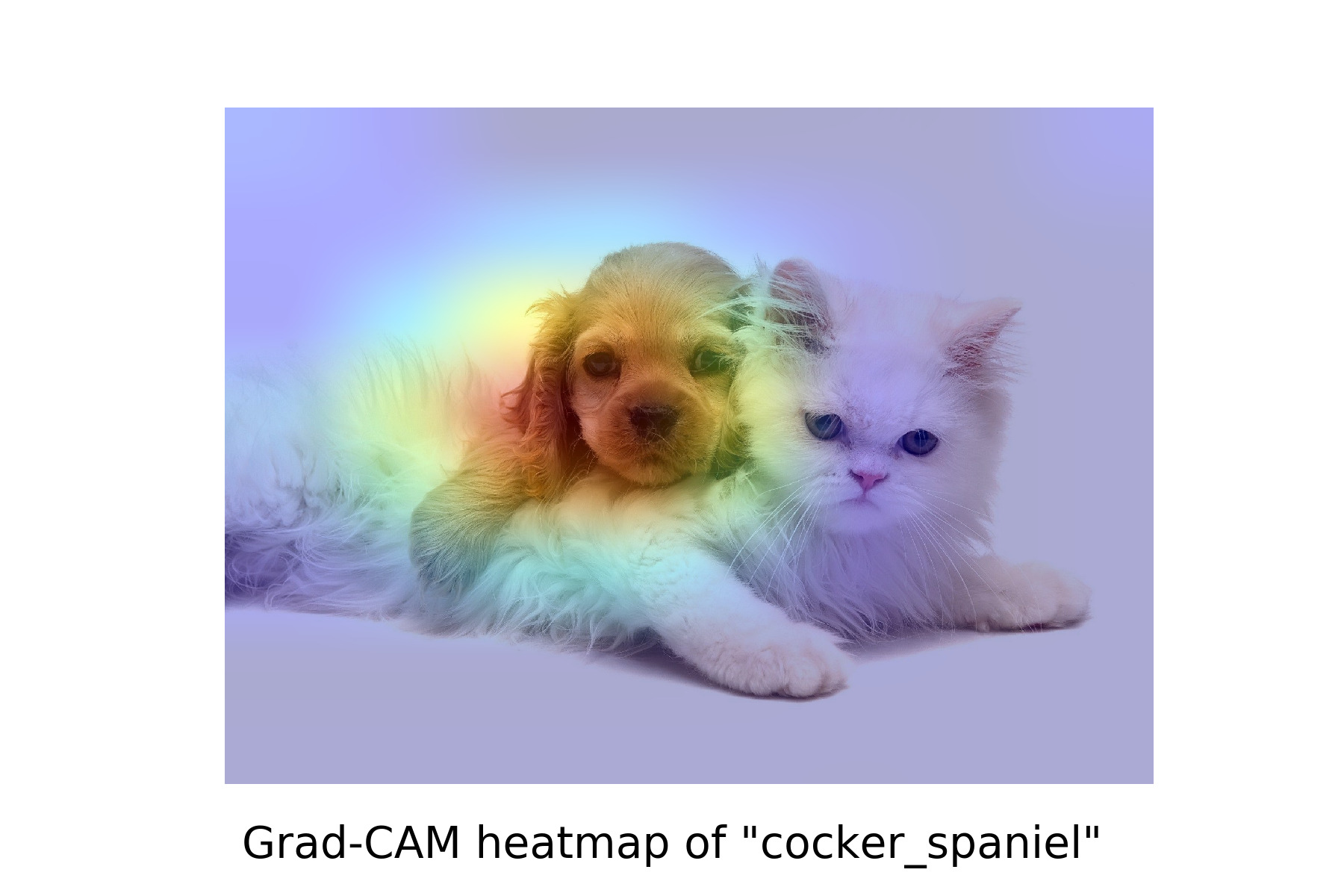} &
    \includegraphics[width=0.32\linewidth, viewport=70 10 370 240, clip]{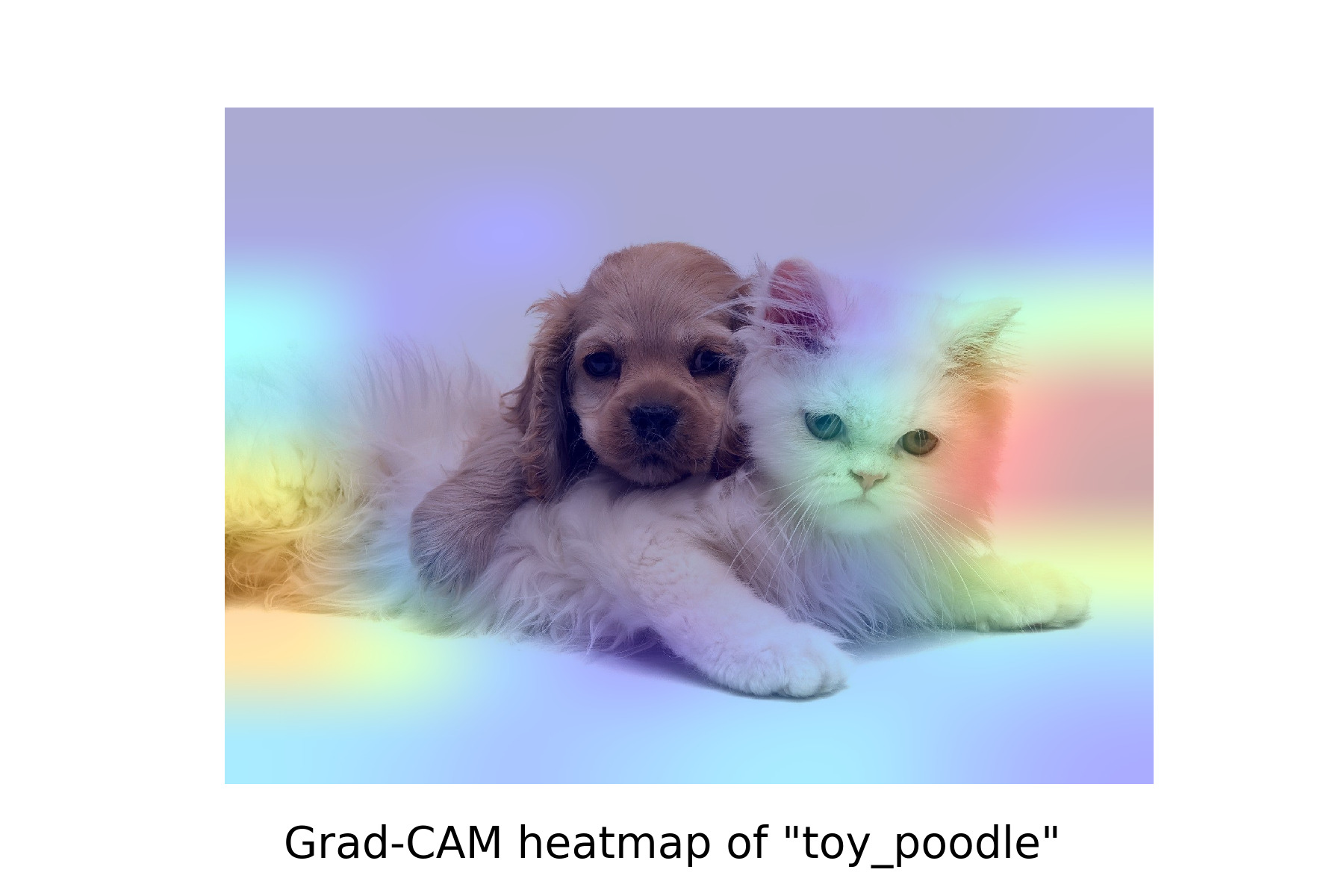} &
    \includegraphics[width=0.32\linewidth, viewport=70 10 370 240, clip]{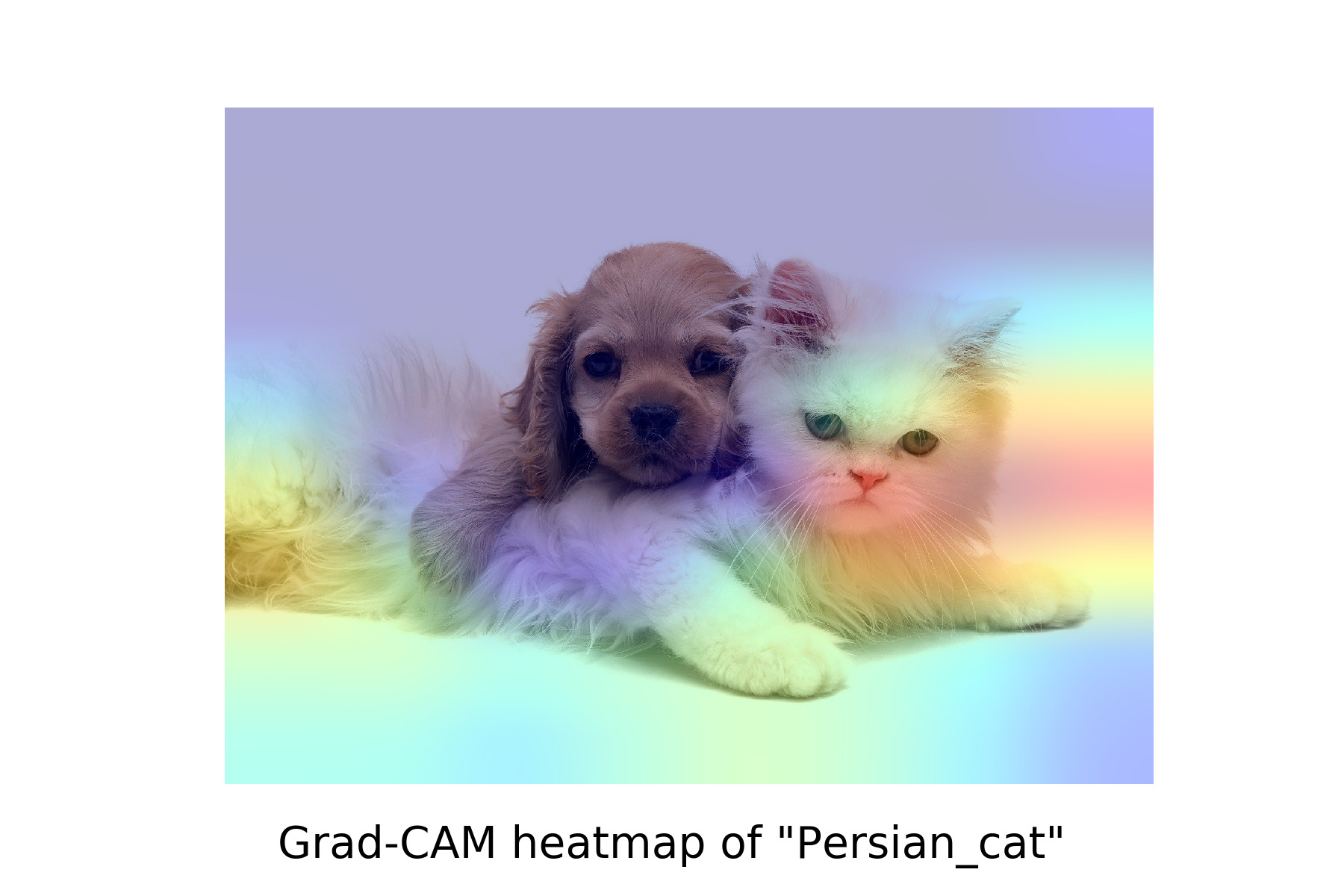} \\
    {\small (a)} & {\small (b)} & {\small (c)} \\
  \end{tabular}
  \caption{\label{fig:ToyPPersCnoLop}\emph{spaniel-kitty-paws-cut}: Grad-CAM explanation of Inception V3 for the identification of the ``cocker spaniel'' (a), the ``toy poodle'' (b) and the ``Persian cat'' (c), when the paws are removed from the image (compare results of Fig.~\ref{fig:inc-spanielPcat}).}
\end{figure}

A careful observation of the delivered network explanations shows that they are partly
arbitrary and hardly intuitive, and this independently of a wrong, or right class prediction.  For example, the network reasoning behind the ``toy poodle'' classification in
Fig.~\ref{fig:inc-spanielPcat} (b), which is wrong as far as the race
of the dog is concerned, but right as far as the animal category
identified (a dog), cannot be noted as sound. The main reason is because the
most activated, and therefore the most relevant to the target
identification region (marked red), points to a part of the image that
lies in empty space, beyond the contour of the target. The marked red region lies close to
what one could describe as a \emph{generic feature}, the paws, which is common to a
variety of animals.
A too \emph{generic feature} offers little confidence in being a good explanation, if assumed that it is only the accuracy of the feature's
localization in the image that fails. Besides, the algorithm could have focused
on the vicinity of the paws out of reasons not directly associated with the recognition of the ``poodle''. Observing that the explanation for the identification of the ``Persian cat'', see Fig.~\ref{fig:inc-spanielPcat} (c), highlights the same paws, makes the unambiguity or definiteness of the explanations questionable.
Important is also the investigation of the
\emph{stability} and \emph{consistency} of the network's explanations, as
they relate to the reproducibility of the network too.
For example, it would be expected that a network which
concentrated on the dog's head to explain the first place of
the top-predictions-list, the ``cocker spaniel'' in  Fig.~\ref{fig:inc-spanielPcat} (a), would probably also pick the head
to mainly identify the second most probable classification on the list, which
the ``toy poodle'', seen in Fig.~\ref{fig:inc-spanielPcat} (b). This is however not the case, which makes the consistency behind the logic of explanations doubtful.
Obviously, the cat's head also receives hardly any attention for the explanation of
the recognition of the cat in \ref{fig:inc-spanielPcat} (c). It is not possible to identify some certain strategy which the network consistently employs in order to explain classifications, in this case of animals.
For further investigations, a small part of the image
\emph{spaniel-kitty}, namely the part containing the paws, has been
removed from the image and the top-predictions-list has been calculated
again. With the new test image, \emph{spaniel-kitty-paws-cut} as
input, the ``cocker spaniel'' keeps the first place on the
top-predictions-list, see Table~\ref{tb:inseption-scores1}, however the
``Persian cat'' climbes now from place 16 to place 2 with a classification
probability rising from 0.5\,\% to 30\,\%, while the ``toy poodle'' falls down to the place 4
of the list.
In Table~\ref{tb:inseption-scores1}, the new top-four predicted classes and their new scores are displayed. There are no great changes in the explanation
concerning the ``cocker spaniel'' for the modified image, the head being the part highlighted again.
However the visual explanations for the identification of the ``toy
poodle'' and the ``cat'' have changed considerably, as in Fig.~\ref{fig:ToyPPersCnoLop} to see.
%
The ``toy poodle'' is now overlayed by a double heat spot, a minor one at the end of the cat's body and the main one to the right of the cat's head, both lying outside the contour of the recognized ``poodle'', see Fig.~\ref{fig:ToyPPersCnoLop}(b). Although in this case the classification is correct, the explanation doesn't make sense at all, because
the activation region lies entirely outside the target (``toy poodle'').
One could argue that at least the explanation for the ``Persian cat'' in Fig.~\ref{fig:ToyPPersCnoLop} (c) has been improved, in comparison to the unchanged image. The hot activation
region approaches now the cat's head instead of the paws which is more characteristic of the target. However, a considerable part of the class activation mapping (marked red), still lies beyond the contour of
the cat and therefore, at least the position of the recognized target, can be described as not accurate or even wrong. Inception V3 delivers identical results, with respect to changing execution environments, therefore the explanations and classifications of the network are
proved to be deterministic under laboratory conditions, that is when no intentional or unintentional perturbations are inserted to the test data. 
\subsection{Xception}
\label{s:xceptionv3exam}
\begin{figure}[htbp]\setlength{\tabcolsep}{0.7mm}
  \centering
  \begin{tabular}{cc}
    \includegraphics[width=0.48\linewidth]{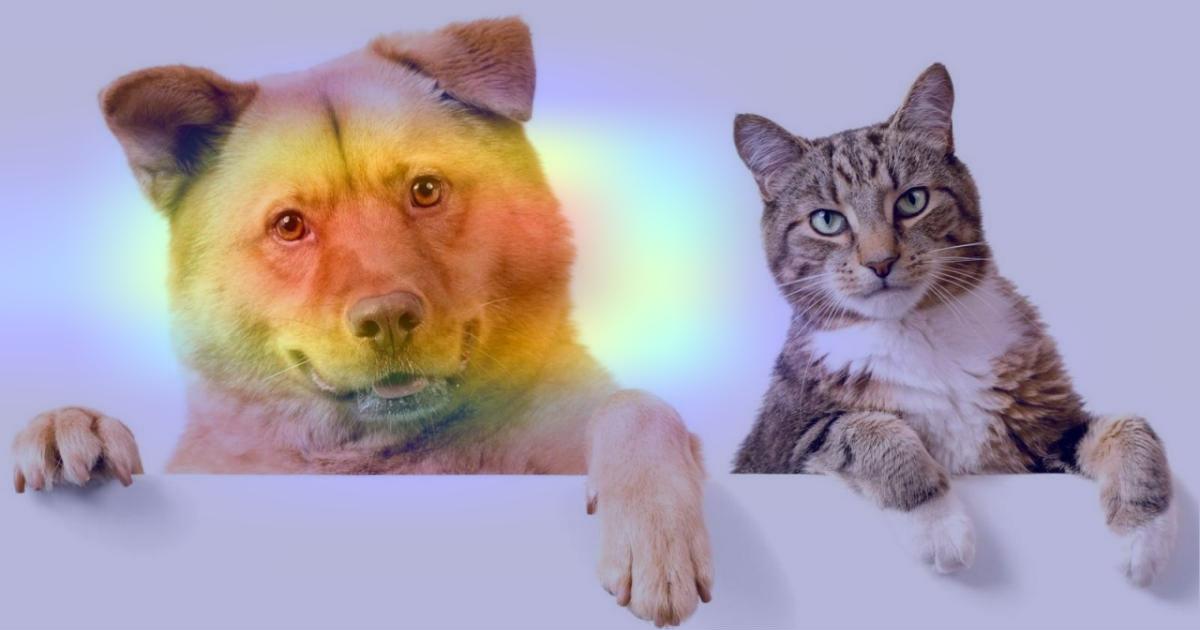} &
    \includegraphics[width=0.48\linewidth]{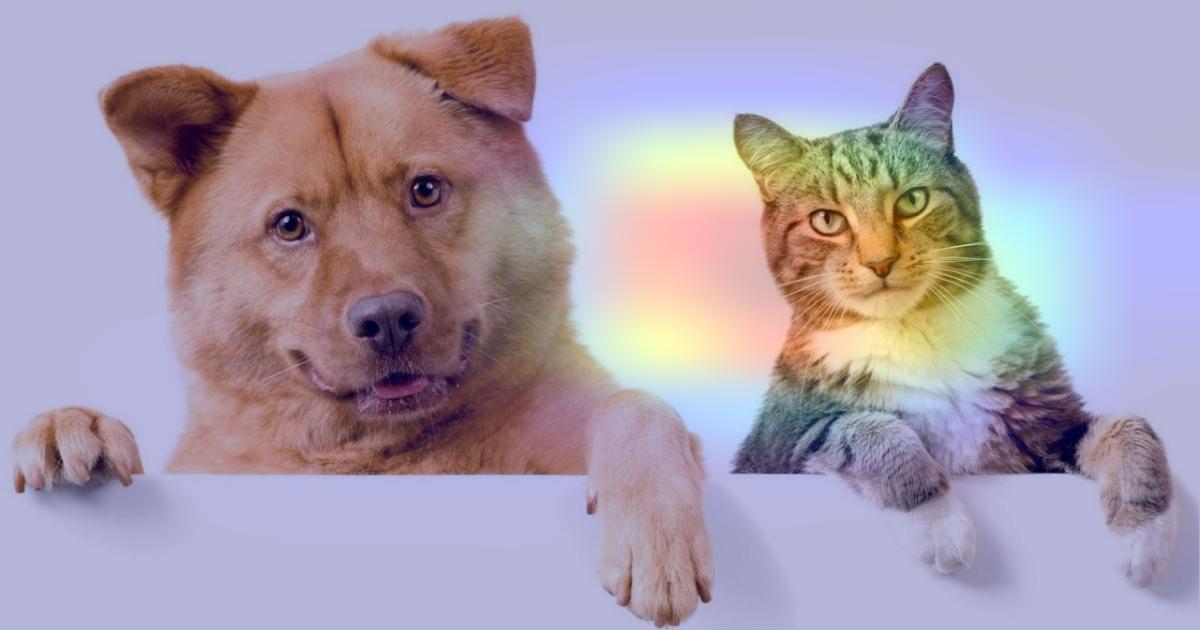} \\
    {\small (a)} \textsf{\footnotesize ''chow''} & {\small (b)} \textsf{\footnotesize ''Egyptian\_cat''} \\
  \end{tabular}
  \caption{\label{fig:chow-cat-x}\emph{chow-cat}: Grad-CAM explanations of Xception for the identification of the ``chow'' (a), in the first place of the top-predictions-list and the ``Egyptian cat'' (b), in the second place. Third on the list is the ``tiger cat'' and fourth the ``tabby cat''. For a comparison, the order of explanations generated by Inception V3 is given in the caption of Fig.~\ref{fig:inc-chowtabby}.}
\end{figure}

In analogy to \ref{s:inceptionv3exam}, object detections and their explanations calculated with the Xception network are here discussed.
In Fig.~\ref{fig:chow-cat-x}(a) and (b) there are presented the activation heatmaps, produced by the network
for the identification of the image regions that
correspond to the ``dog'' (``chow''), and the ``cat'' respectively, (here identified as ``Egyptian cat'', whereas Inception V3 identified the cat as a ``Tabby cat'', compare Fig.\ref{fig:inc-chowtabby}).
%

\begin{figure}[htbp]\setlength{\tabcolsep}{0.5mm}
  \centering
  \begin{tabular}{cc}
    \includegraphics[width=0.48\linewidth]{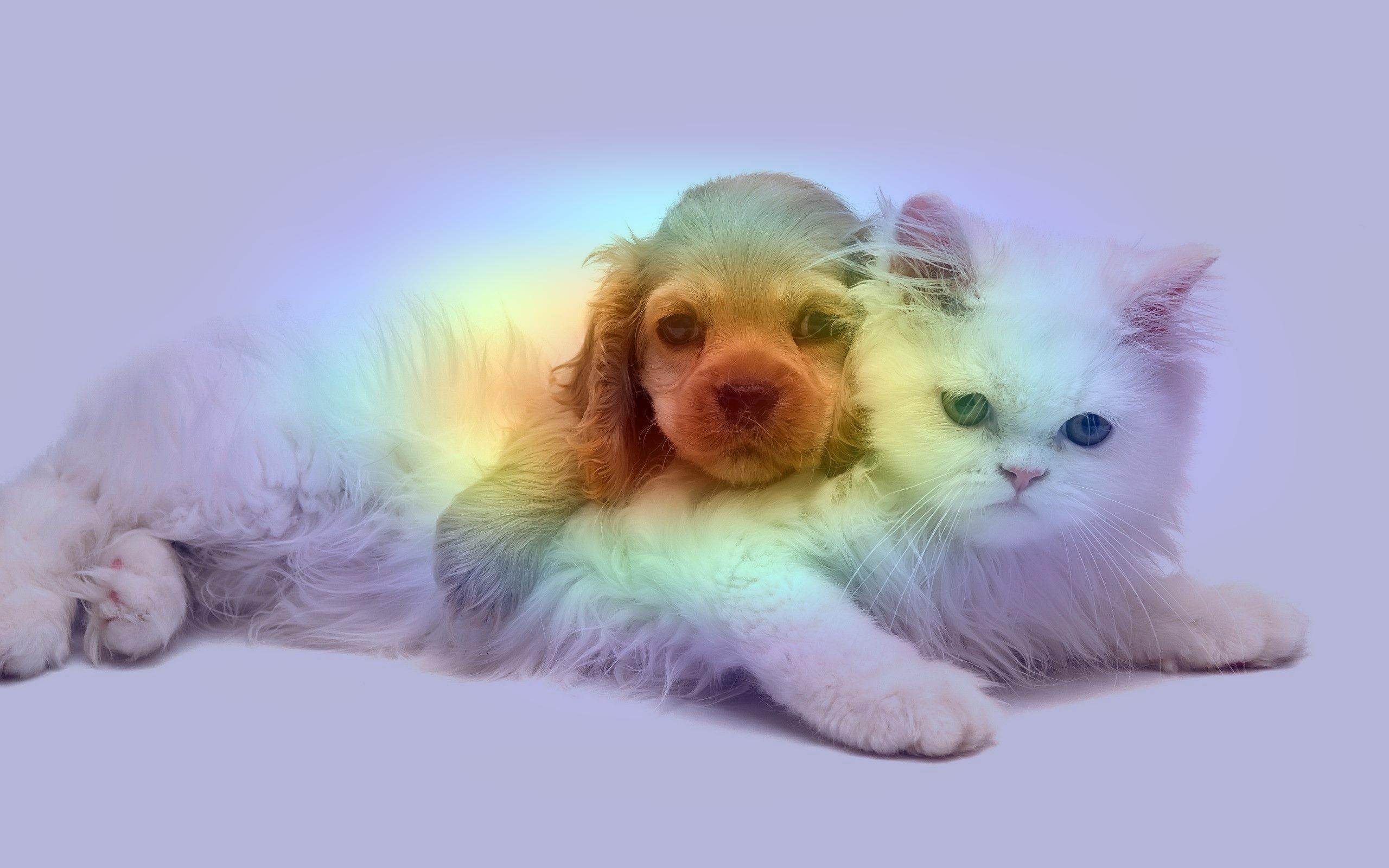} &
    \includegraphics[width=0.48\linewidth]{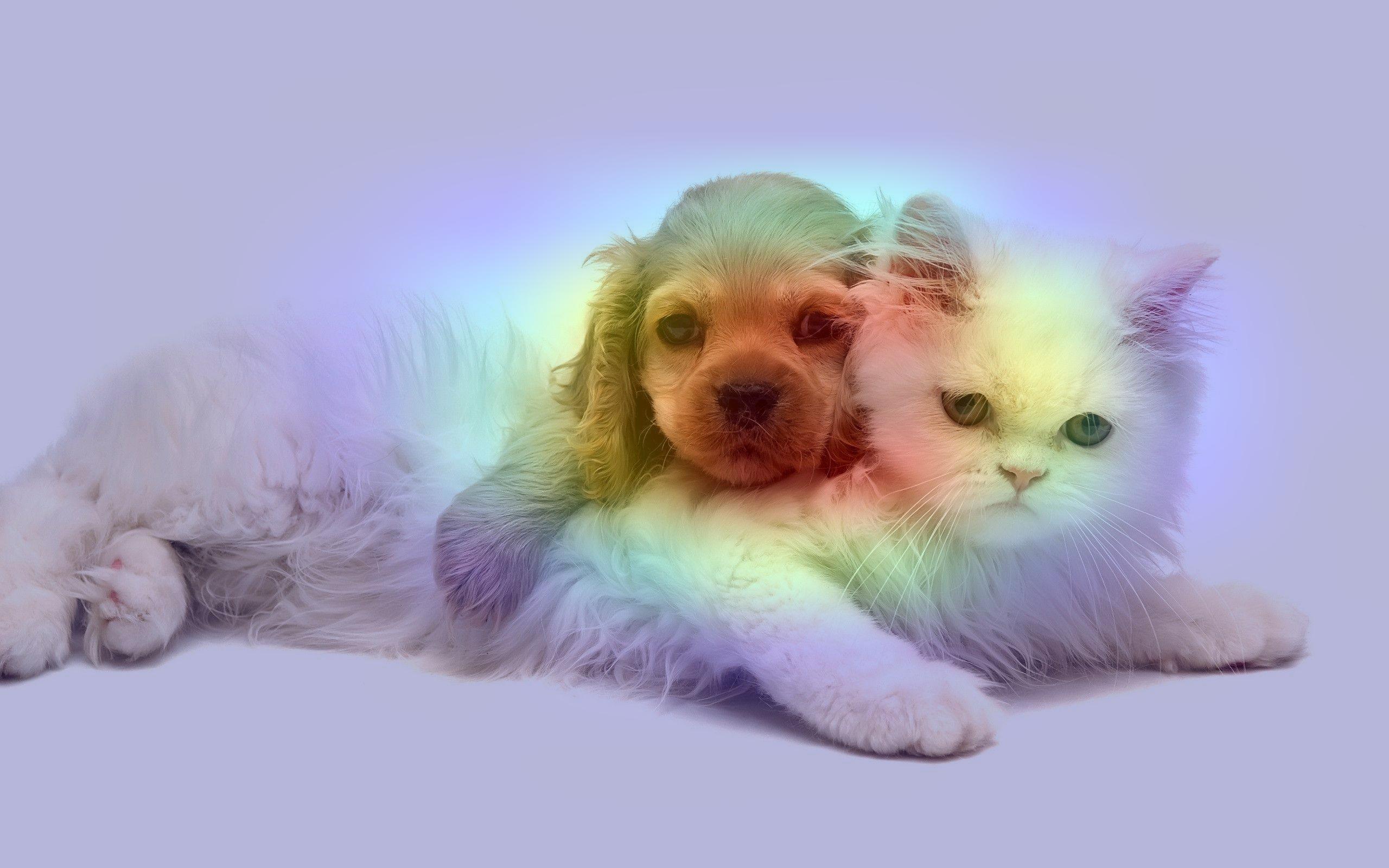} \\
    {\small (a)} \textsf{\footnotesize ''cocker\_spaniel''} & {\small (b)} \textsf{\footnotesize ''French\_bulldog''} \\
   \includegraphics[width=0.48\linewidth]{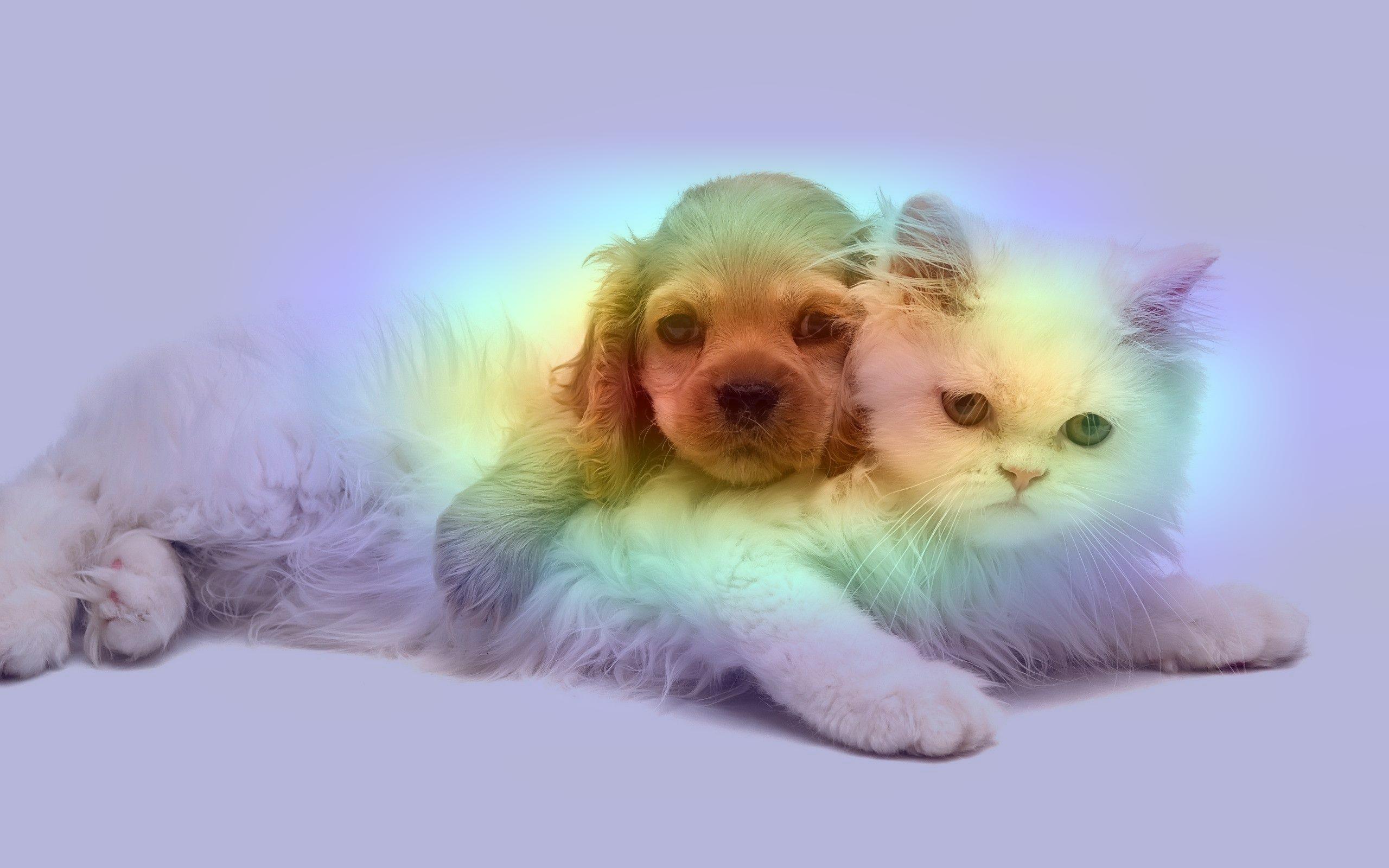} &
    \includegraphics[width=0.48\linewidth]{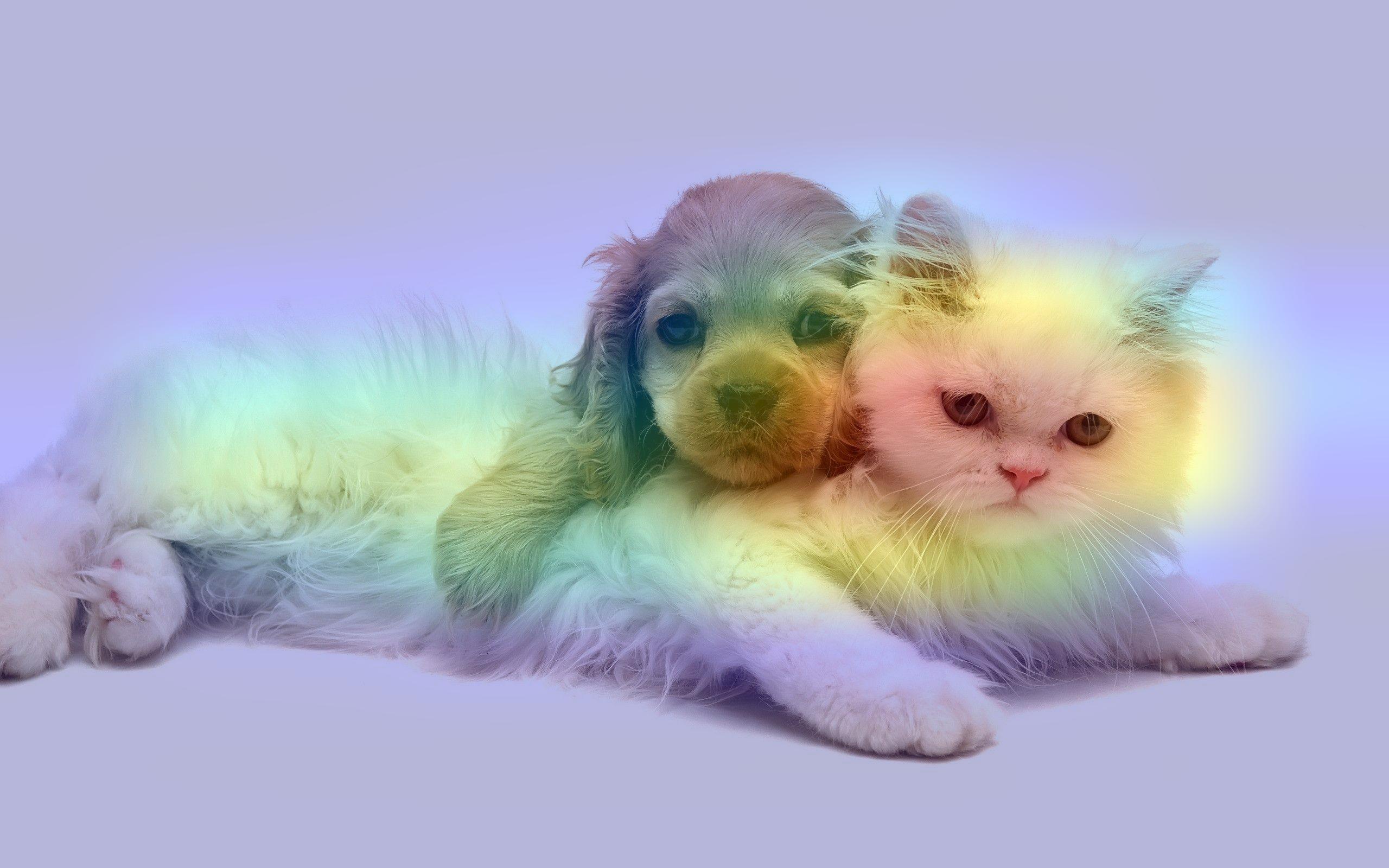} \\
    {\small (c)} \textsf{\footnotesize ''toy poodle''} & {\small (d)} \textsf{\footnotesize ''Persian cat''} \\
  \end{tabular}
  \caption{\label{fig:spaniel-bulldogx}\emph{spaniel-kitty}: Grad-CAM explanations of Xception for the identification of the ``cocker spaniel'' (a), the ``French bulldog'' (b), the ``toy poodle'' (c) and the ``Persian cat'' (d).}
\end{figure}

\begin{figure}[htbp]\setlength{\tabcolsep}{0.7mm}
  \centering
  \begin{tabular}{ccc}
    \includegraphics[width=0.33\linewidth]{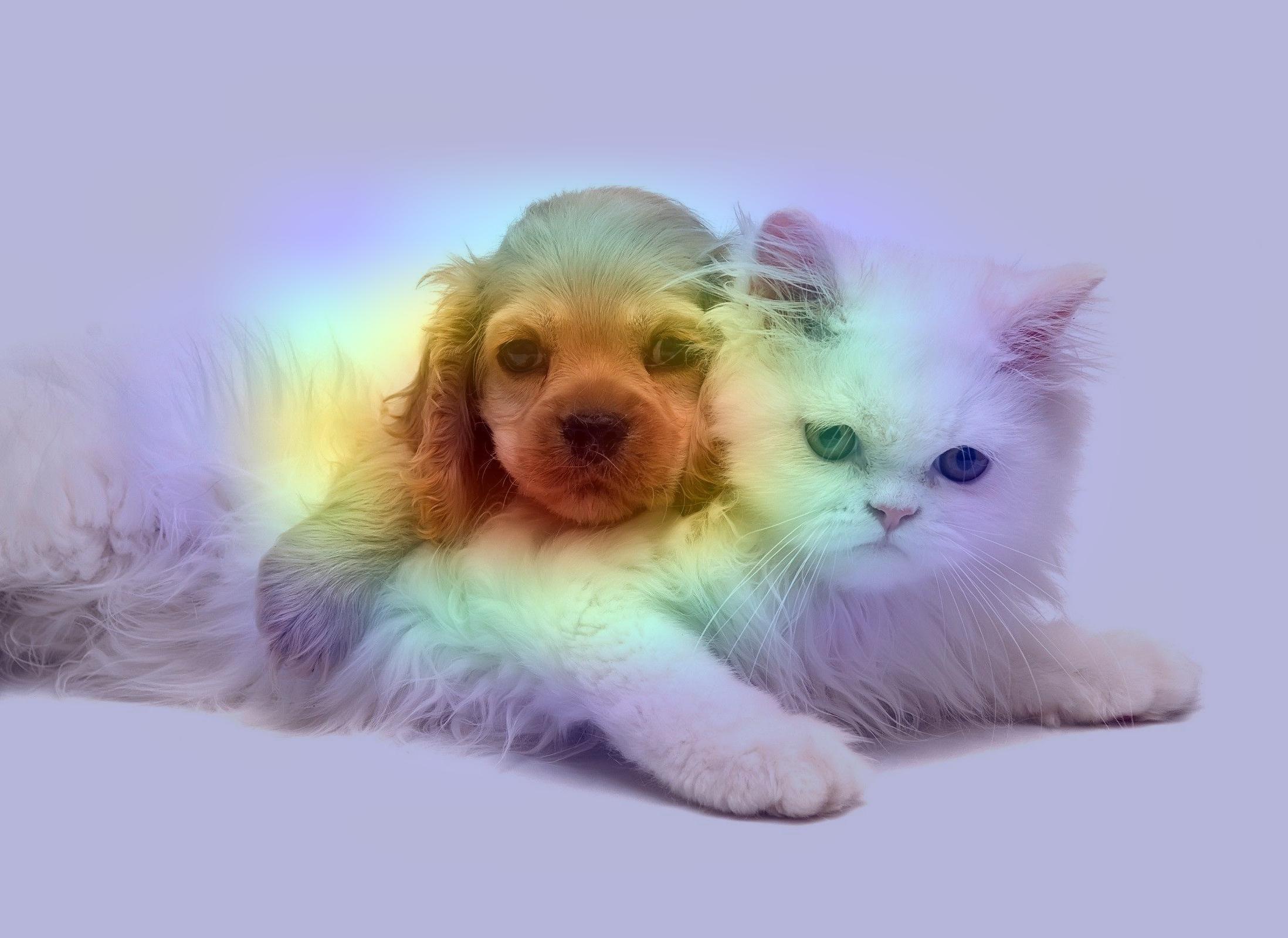} &
    \includegraphics[width=0.33\linewidth]{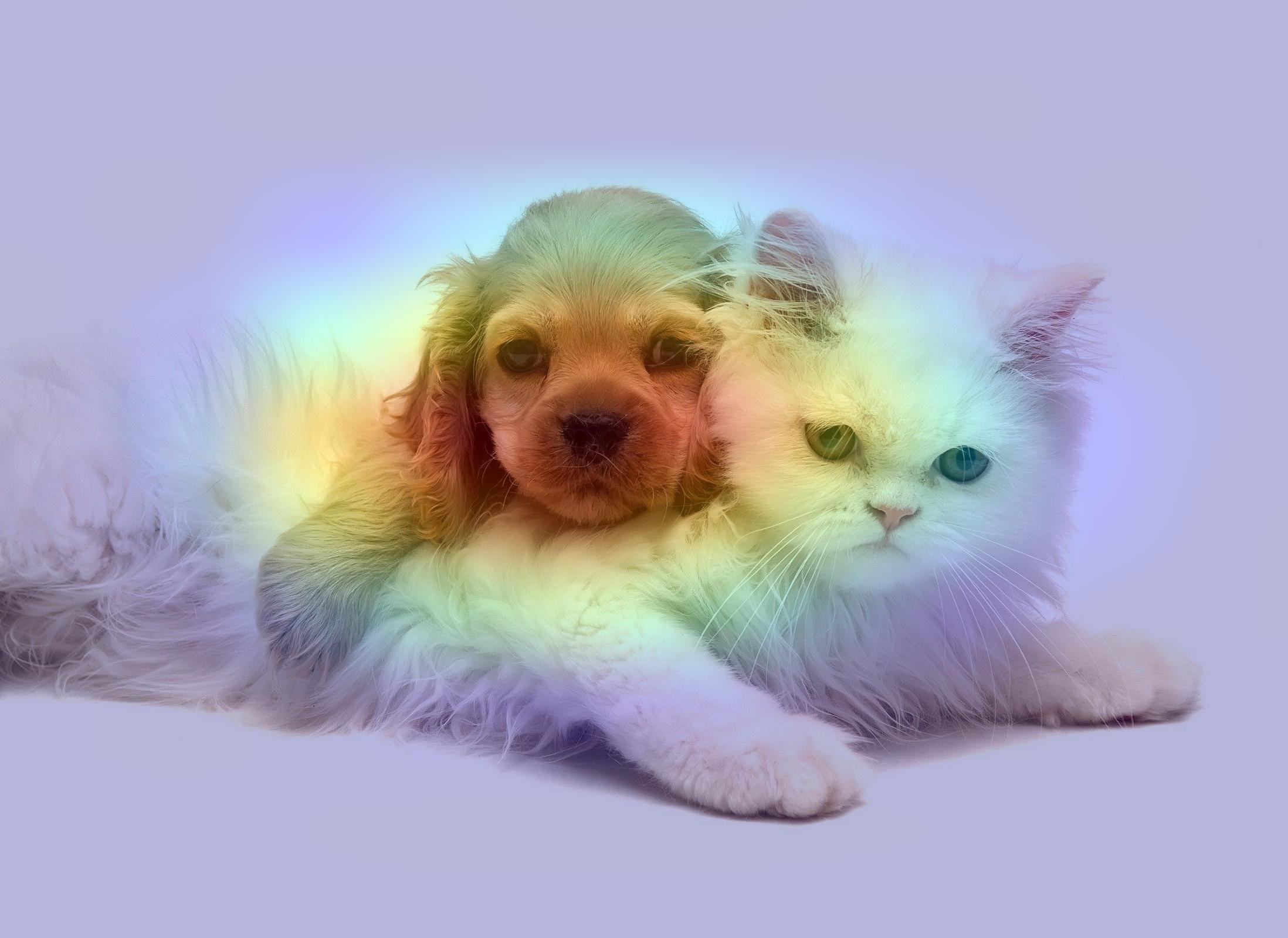} &
    \includegraphics[width=0.33\linewidth]{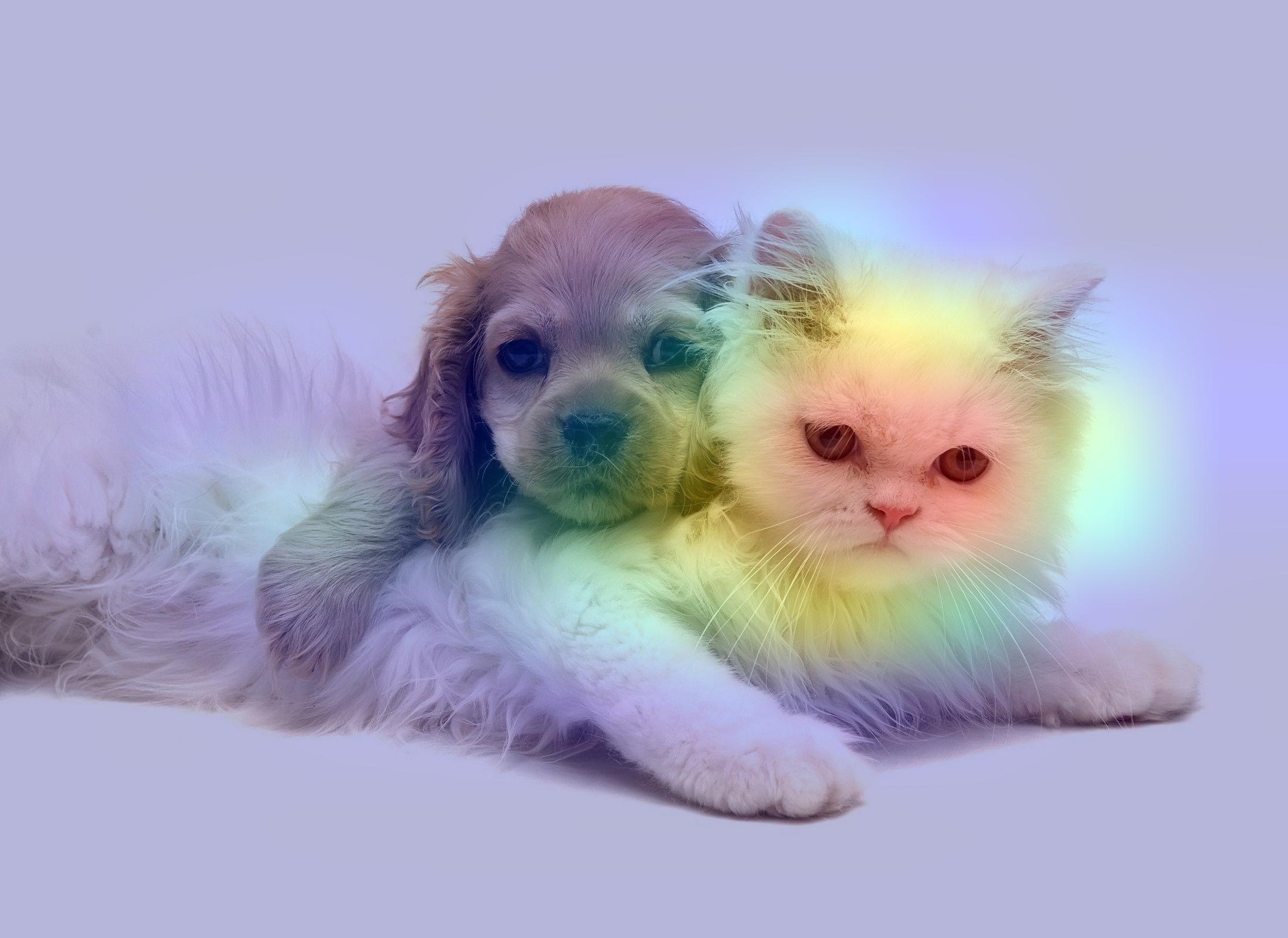} \\
    {\small (a)} \textsf{\footnotesize ''cocker spaniel''} & {\small (b)} \textsf{\footnotesize ''toy\_poodle''} & {\small (c)} \textsf{\footnotesize ''Persian cat''}\\
  \end{tabular}
  \caption{\label{fig:ToyPersianCat-x-cut}\emph{spaniel-kitty-paws-cut}: Grad-CAM explanations of Xception for the identification of the ``cocker spaniel'' (a), the ``toy poodle'' (b) and the ``Persian cat'' (c) with paws removed from the image (compare Fig.~\ref{fig:spaniel-bulldogx}).}
\end{figure}

In Fig.~\ref{fig:spaniel-bulldogx} the activation maps
corresponding to the identification of the ``cocker spaniel'', the
``French bulldog'', the ``toy poodle'' and the ``Persian cat'' respectively are demonstrated.
Similarly to the Inception V3 case, described in the previous
section, all prediction scores are almost identical between all
HW environment executions.
Again, to check the stability of the visual explanations, the identifications of the ``cocker spaniel'', the ``toy poodle'' and the ``Persian cat'' on the image \emph{spaniel-kitty-paws-cut} have also been tested and the results are depicted in Fig.~\ref{fig:ToyPersianCat-x-cut}.
%
%
Xception seems to consistently focus on the head of the identified
animal for the classification and also doesn't produce activation
regions outside the contour of the target. The consistency of feature attribution for the explanation and  the accuracy of
the target position are thus better for Xception in comparison to
Inception V3.  Because the removal of a part of the image (paws) didn't
affect the explanations, at least for the here described experiments, the explanations of Xception can be described as more stable in comparison to the explanations of Inception V3.
As far as the definiteness of the classifications and the stability of their scores are concerned, there are following remarks: Xception can unambigously and correctly distinguish between different classes, focusing on the respective region of the image that depicts the accordingly recognized class. However the differentiation of subclasses within a class, here for example of dog breeds within the class \emph{dogs}, is neither definite nor stable.
For instance, all identifications of the dog races listed in Table~\ref{tb:Xception-scores}, produce almost one and the same visual explanation which is close to identical to the explanation for the ``cocker spaniel'', on the first place of the top-predictions-list, (see Figs.~\ref{fig:ToyPersianCat-x-cut} and~\ref{fig:spaniel-bulldogx}).
This problem probably relates to the quality of \emph{image segmentation} and
\emph{fine-grained classification} commonly  arising when categories share a number of similar attributes which are also contained within the same bounding box.
%
\begin{table}[htbp]
  \caption{\label{tb:Xception-scores}Xception top-predictions-list for the images \emph{spaniel-kitty} (A) and \emph{spaniel-kitty-paws-cut} (B).}
  \begin{minipage}[t]{\linewidth}
  \centering
  \vspace{2mm}
  \begin{tabular}{| l |c | c |}
  \hline
               &  A & B \\
  \hline
  1 &   Cocker spaniel        &  Cocker spaniel \\
  2 &     clumber       &   toy poodle \\
  3 &      Sussex spaniel         &  Blenheim spaniel \\
  4 &      Blenheim spaniel       &   Pekinese \\
  5 &      Golden retriever         & Chihuahua \\
  6 &     Pekinese               &  Sussex spaniel\\
  7 &     Chow               &  clumber \\
  8 &    Toy poodle             &  Golden retriever\\
  9 &    Chihuahua              & miniature poodle \\
  10 &   French bulldog       &  Chow \\
  11 &   Persian cat           & Pomeranian \\
  12 &   Shih-Tzu           & Shih-Tzu \\
  13 &   Pomeranian           & Persian cat \\
  \hline
  \end{tabular}
  \end{minipage}%
\end{table}
\subsection{Inception V3 vs. Xception}
\begin{figure}[htbp]\setlength{\tabcolsep}{0.5mm}
  \centering
  \begin{tabular}{ccc}
    \includegraphics[width=0.32\linewidth, viewport=120 34 330 250, clip]{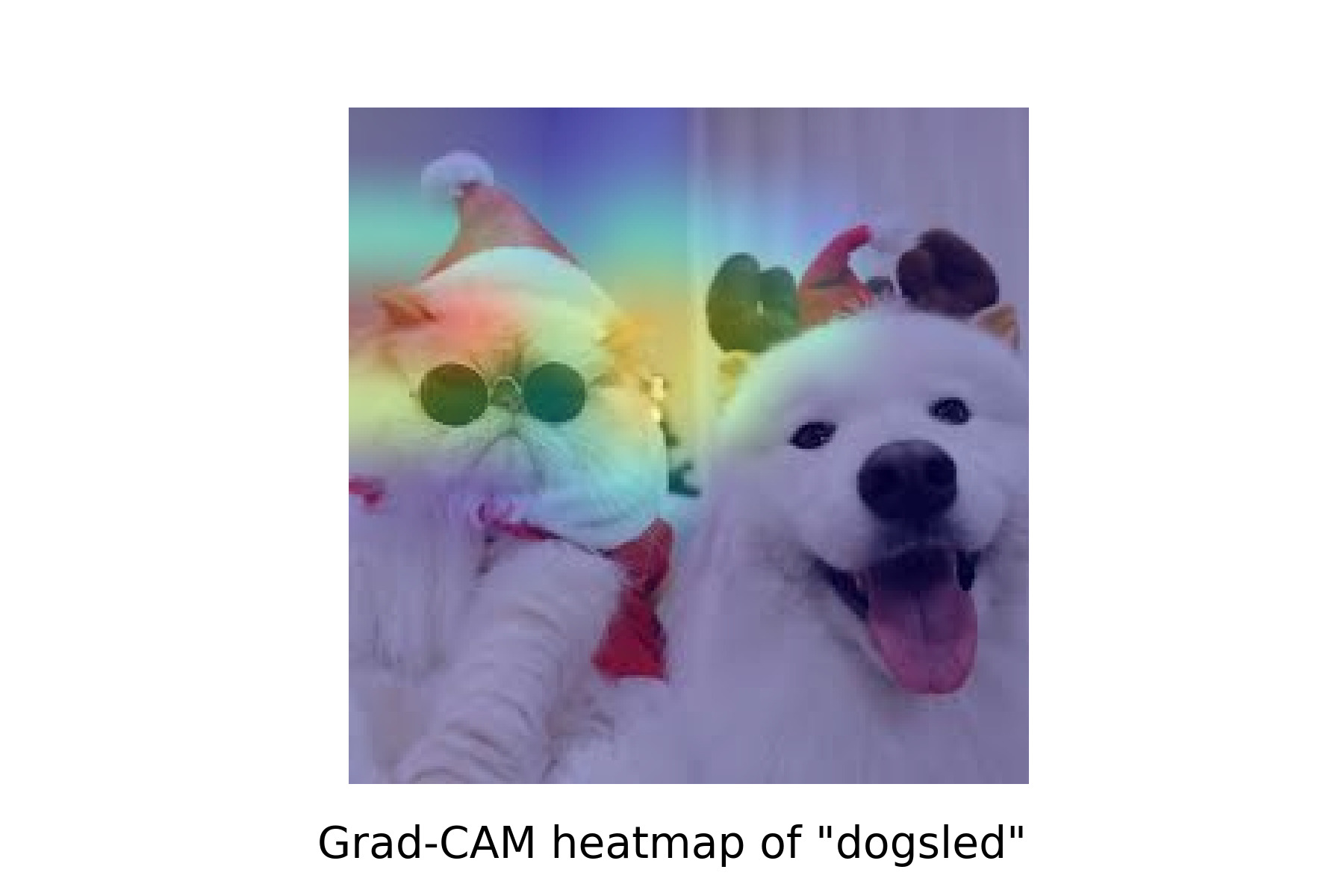} &
    \includegraphics[width=0.32\linewidth, viewport=120 34 330 250, clip]{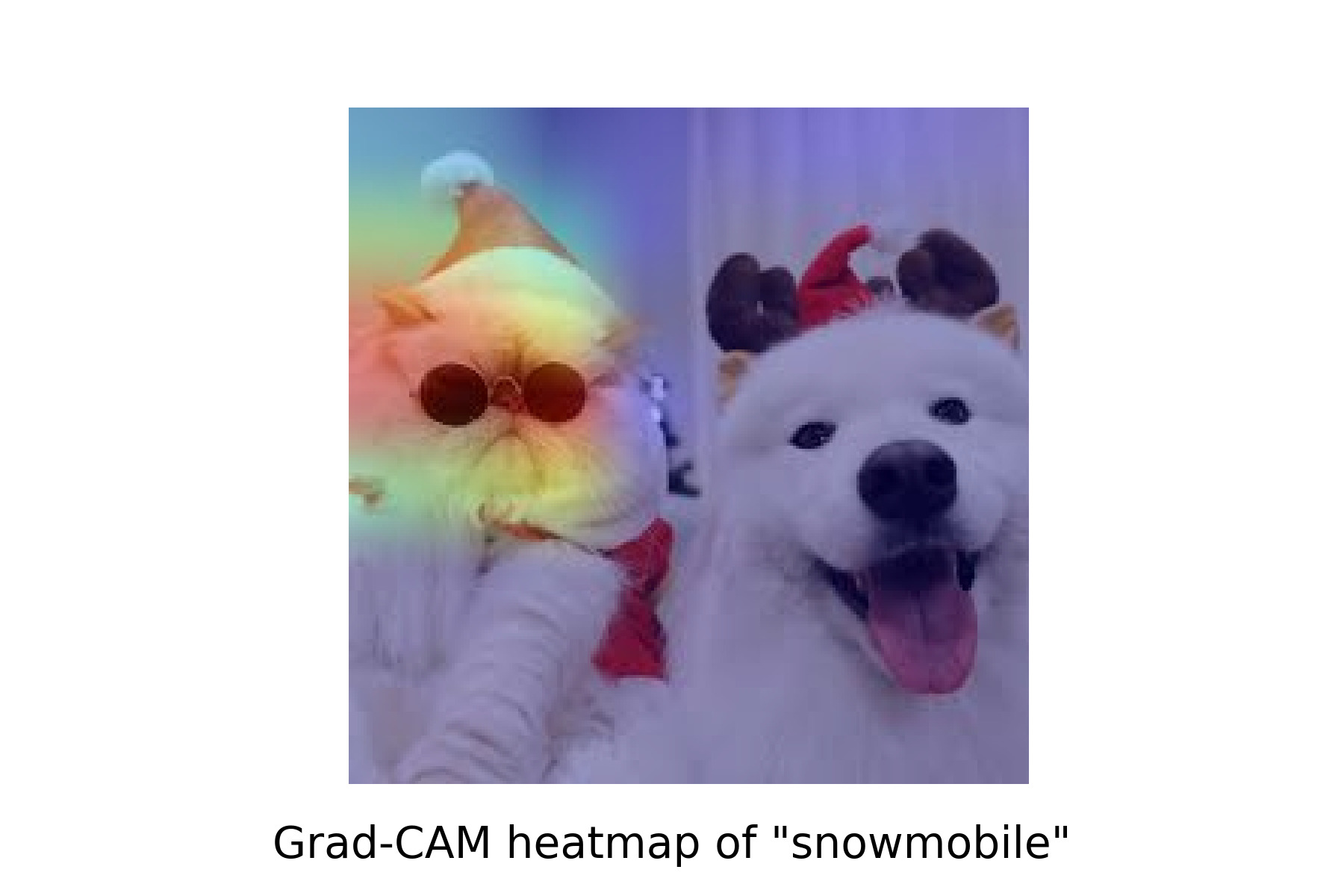} &
    \includegraphics[width=0.32\linewidth, viewport=120 34 330 250, clip]{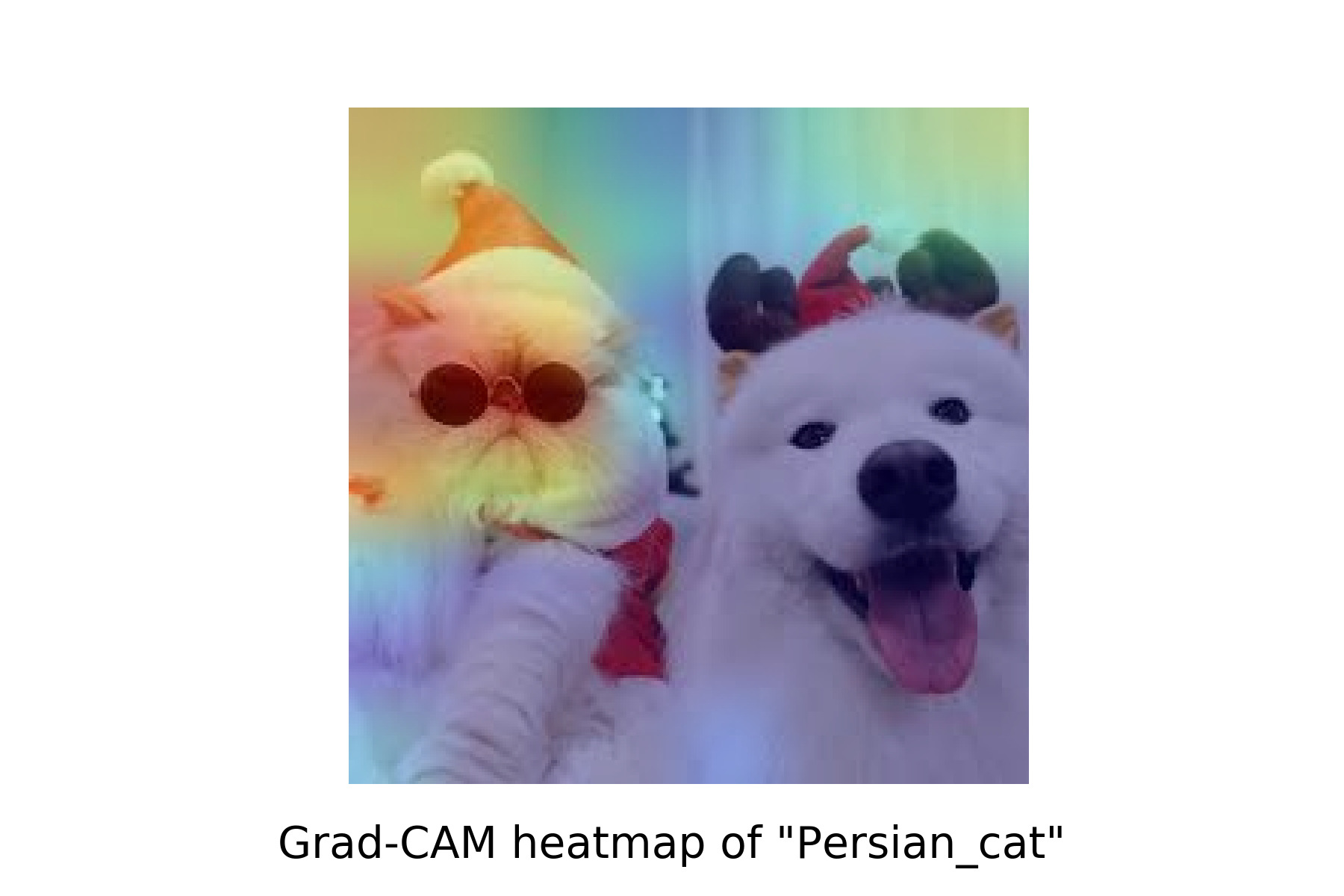} \\
    {\small (a)} \textsf{\footnotesize ''dog sled''} & {\small (b)} \textsf{\footnotesize ''snowmobile''} & {\small (c)} \textsf{\footnotesize ''Persian cat''} \\
  \end{tabular}
  \caption{\label{fig:xmas-cat-dog-inc}\emph{xmas-cat-dog}: Grad-CAM explanations of Inception V3 for the identification of a ``dog sled'' (a), a ``snowmobile'' (b) and a ``Persian cat'' (c). The classification probability decreases from (a) to (c).}
\end{figure}

\begin{figure}[htbp]\setlength{\tabcolsep}{0.7mm}
  \centering
  \begin{tabular}{cc}
    \includegraphics[width=0.40\linewidth]{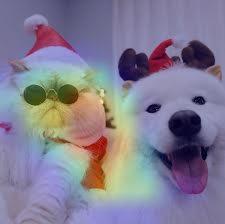} &
    \includegraphics[width=0.40\linewidth]{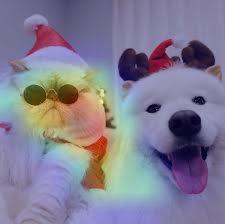} \\
    {\small (a)} \textsf{\footnotesize  ''Angora\_cat''} & {\small (b)} \textsf{\footnotesize ''Persian\_cat''} \\
  \end{tabular}
  \caption{\label{fig:xmas-cat-dog-x-cats}\emph{xmas-cat-dog}: Grad-CAM explanations of Xception for the identification of an ``Angora cat'' (a) and a ``Persian cat'' (b).}
\end{figure}

\begin{figure}[htbp]\setlength{\tabcolsep}{0.7mm}
  \centering
  \begin{tabular}{ccc}
    \includegraphics[width=0.32\linewidth]{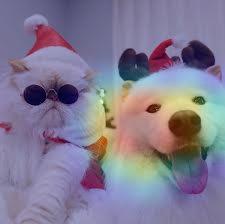} &
    \includegraphics[width=0.32\linewidth]{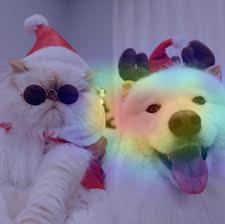} &
    \includegraphics[width=0.32\linewidth]{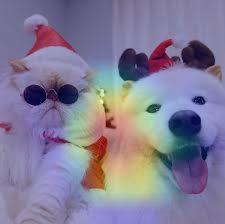} \\
    {\small (a)} \textsf{\footnotesize ''Samoyed'' dog} & {\small (b)} \textsf{\footnotesize ''dog sled''} & {\small (c)} \textsf{\footnotesize ''Maltese\_dog''} \\
  \end{tabular}
  \caption{\label{fig:xmas-cat-dog-x-cats-maltese}\emph{xmas-cat-dog}: Grad-CAM explanations of Xception for the identification of ``Samoyed'' (a), a ``dog sled'' (b), and a ``Maltese dog'' (c). The classification probability decreases from (a) to (c).}
\end{figure}

Explanation properties like correctness, stability, definiteness and accuracy have been already indicated while discussing the images in the previous sections. As already mentioned, Xception appears to have certain
advantages in comparison to Inception V3.
However, further experiments with both networks, convince that a deeper evaluation should be necessary, if these networks have to be trusted for critical applications.
For example in Fig.~\ref{fig:xmas-cat-dog-inc}, Inception V3 has identified the ``Persian cat''
at the seventeeth place on the top-predictions-list but it has also identified a ``snowmobile'' with a higher probability than that of the cat and a ``dog sled'', the latter at the fourth from the top position on the top-predictions-list.
The activated image region to explain
the ``Persian cat'' is remarkably similar to the one which explains the ``snowmobile''. Puzzling is that for the explanation of the ``Persian cat'', the network highlights also a completely irrelevant region in the background of the image, to the right over the head of the ``Samoyed dog''. Definiteness and correctness of the explanations grow rapidly weaker with growing distance from the first score on the top-predictions-list. This well-known problem is obviously critical, especially in case it is necessary to simultaneously classify more than one object categories in one image. It should be stressed, that the here discussed identifications concern objects that belong to the first top-seventeen out of one thousand classes that the network can recognize.
The activation regions produced with Inception V3 for the ``Persian cat'', the ``snowmobile'' and the ``dog sled'' are to their greatest part overlapping, see Fig.~\ref{fig:xmas-cat-dog-inc}.
In Fig.~\ref{fig:xmas-cat-dog-x-cats}, Xception's consistency with regard to the choice of attributes to recognize members of the same class, is demonstrated. It seems to remain robust, also when lower rank scores are explained, as this example of identification of cats shows, with the ``Angora'' at the twentieth and the ``Persian'' at the twenty-sixth positions on the top-predictions-list.
In Fig.~\ref{fig:xmas-cat-dog-x-cats-maltese}, Xception highlights image regions that correspond to the classification of the ``Samoyed'' (first position), ``dog sled'' (ninth position) and ``Maltese dog'' (fourteenth position) on the top-predictions-list.
One can see, that the visual activation for the ``Samoyed'' (a) has much in common with that for the ``dog sled'' (b), while the explanation for the ``Maltese dog'', placed in the gap between the ``Samoyed'' and the ``Persian cat'', is obviously wrong
at least as far as the accuracy of the position of the identified object is concerned and indefinite as regards the class of the object. It is obvious that the consistency of attributes makes hardly any sense without the definiteness of classifications.

\section{Self-trained Models}
\label{sec:DetSelfTrainMod}
The impact of various HW architectures especially on model training will be here investigated after making the source code as deterministic as possible.
For the training of the here discussed models, the HW execution environments listed at the end of \ref{s:archHW} have been employed.
%
%
\subsection{Deterministic ConvNet}
A simple convolutional network (ConvNet) created with \lsti|TensorFlow 2| and \lsti|Keras| was trained as binary classifier for pictures showing cats and dogs.
It consists of alternated \lsti|Conv2D| (with relu activation) and \lsti|MaxPooling2D| layers with five convolutional and four \lsti|Pooling| layers,
plus an input and a pre-processing layer,
as well as a \lsti|Flatten| and a so called \lsti|Dense| layer with \emph{sigmoid} activation which ends
the model. Flattening is necessary to transform 3D tensor outputs to 1D inputs for the \lsti|Dense| layer.
The network is described in~\cref{l:ConvNetcode}. The image dimension in all test models, \lsti|PDIM|, is set to 180.
The training data have been taken from the dataset \emph{dogs-vs-cats}
provided by a Kaggle competition~\cite{DogsCats}.
A limited number of altogether 2,000 images has been used for training and 1,000 for the validation of the network, with equal number of cat and dog images.
Binary cross entropy was used as loss function for the training together with the rmsprop optimizer in its default configuration and as metrics was chosen accuracy.
The main task of the network was to create reproducible results, that is to create a deterministic training process. To this purpose three
OS environment variables had to be set first. Using a Jupyter notebook this can be accomplished by a special Jupyter kernel. The setting of the variables reads:
\begin{lstcode}[language=sh,caption={\label{l:EnvVars}Environment variables set for determinism.}]
  PYTHONHASHSEED=1
  TF_DETERMINISTIC_OPS=1
  TF_CUDNN_DETERMINISTIC=1 
\end{lstcode}
In the Python code several pseudo-random generators have to be initialized
with a \emph{fixed seed}, Python, NumPy, and TensorFlow RNGs are seeded.
The according code is as follows:
\begin{lstcode}[caption={\label{l:seeding}Pseudo-RNG seeding set for determinism.}]
  import numpy as np
  import tensorflow as tf

  seed = 1001
  tf.random.set_seed(seed)
  np.random.seed(seed)
  random.seed(seed)
\end{lstcode}
Because a data augmentation and a dropout-layer have been
added to the network to improve accuracy and avoid overfitting, also the seed for
the random operations of \emph{flipping}, \emph{rotating} and \emph{zooming} of data has to be set.
The resulting network has a total of 991,041 trainable parameters.
%
\begin{lstcode}[float=*ht,caption={\label{l:ConvNetcode}Deterministic ConvNet}.]
    data_augmentation = keras.Sequential([kl_exp_preproc.RandomFlip("horizontal", seed=1),
                                          kl_exp_preproc.RandomRotation(0.1, seed=1),
                                          kl_exp_preproc.RandomZoom(0.2, seed=1),])
    PDIM = 180
    inputs = keras.Input(shape=(PDIM, PDIM, 3))
    x = data_augmentation(inputs)
    x = keras.layers.experimental.preprocessing.Rescaling(1./255)(x)
    x = keras.layers.Conv2D(filters=32, kernel_size=3, activation="relu")(x)
    x = keras.layers.MaxPooling2D(pool_size=2)(x)
    x = keras.layers.Conv2D(filters=64, kernel_size=3, activation="relu")(x)
    x = keras.layers.MaxPooling2D(pool_size=2)(x)
    x = keras.layers.Conv2D(filters=128, kernel_size=3, activation="relu")(x)
    x = keras.layers.MaxPooling2D(pool_size=2)(x)
    x = keras.layers.Conv2D(filters=256, kernel_size=3, activation="relu")(x)
    x = keras.layers.MaxPooling2D(pool_size=2)(x)
    x = keras.layers.Conv2D(filters=256, kernel_size=3, activation="relu")(x)
    x = keras.layers.Flatten()(x)
    x = keras.layers.Dropout(0.5, seed=7001)(x)
    outputs = keras.layers.Dense(1, activation="sigmoid")(x)
    model = keras.Model(inputs=inputs, outputs=outputs) 
    model.compile(loss="binary_crossentropy", optimizer="rmsprop", metrics=["accuracy"])
\end{lstcode}
The reproducibility of the training code of \cref{l:ConvNetcode} has been tested and could be confirmed for each execution environment separately, by comparing the resulting parameters and accuracy after the completion of each training process. 
In ~\cref{lst:gpuA} and~\cref{lst:gpuB} the accuracy and a number of
corresponding network weights which belong to the first and to the last but one
layers of the model are displayed, as calculated in the HW-1 and HW-2 environments, respectively, after 90 epochs of training. Although the accuracies in the two execution environments are almost identical, the calculated parameters are completely different, which shows that each environment causes the model to converge to a different optimization minimum.
\begin{lstcode}[float=*ht,caption={\label{lst:gpuA}ConvNet on HW-1: Deterministic training results.}]
  Out: accuracy: 0.849
  
  In : model_weights[0][:,:,0,0]
  Out: <tf.Tensor: shape=(3, 3), dtype=float32, numpy=
       array([[-0.00111067, -0.08925765, -0.03877171],
              [ 0.06882513,  0.07853891, -0.07341643],
              [-0.10082538,  0.05409119, -0.11010181]], dtype=float32)>

  In : model_weights[-2][:32,0]
  Out: <tf.Tensor: shape=(32,), dtype=float32, numpy=
       array([ 0.02049963, -0.06807294,  0.01771718,  0.00662495,  0.09211864,
               0.03866815,  0.05907019,  0.00925133,  0.02810607, -0.00849974,
               0.0199494 ,  0.01444575, -0.07486248, -0.05971878, -0.04888046,
               0.05762889, -0.114383  , -0.03059661, -0.00386733,  0.04686596,
               0.11158869, -0.00510099, -0.01760828,  0.00915093,  0.06470113,
               0.02472718, -0.17019744, -0.06843327,  0.11263403, -0.05122007,
              -0.02696232,  0.01008716], dtype=float32)>
\end{lstcode}
%
\begin{lstcode}[float=*ht,caption={\label{lst:gpuB}ConvNet on HW-2: Deterministic training results.}]
  Out: accuracy: 0.836

  In:  model_weights[0][:,:,0,0]
  Out: <tf.Tensor: shape=(3, 3), dtype=float32, numpy=
       array([[-0.02197945, -0.11502645, -0.06399402],
              [ 0.06629982,  0.07693411, -0.07907421],
              [-0.08929715,  0.07290921, -0.10029382]], dtype=float32)>

  In: model_weights[-2][:32,0]
  Out: <tf.Tensor: shape=(32,), dtype=float32, numpy=
       array([-0.13660763,  0.08144909,  0.0058543 , -0.02693423,  0.02785763,
               0.16917986, -0.00647367, -0.00365615, -0.01310152,  0.04297437,
               0.00301503,  0.01947881, -0.09520608, -0.05354575, -0.06465469,
               0.01503169, -0.06535202, -0.06438911,  0.07274752, -0.01518494,
              -0.04213503,  0.0165184 , -0.04465574, -0.003416  , -0.09923518,
              -0.00701612, -0.05580736,  0.06382342,  0.07347441, -0.06864858,
              -0.05487159, -0.01903648], dtype=float32)>
\end{lstcode}
\subsubsection*{Deterministic ConvNet Summary}
\begin{figure}[htbp]\setlength{\tabcolsep}{0.7mm}
  \centering
  \begin{subfigure}{0.96\linewidth}
    \centering
    \includegraphics[width=0.7\linewidth]{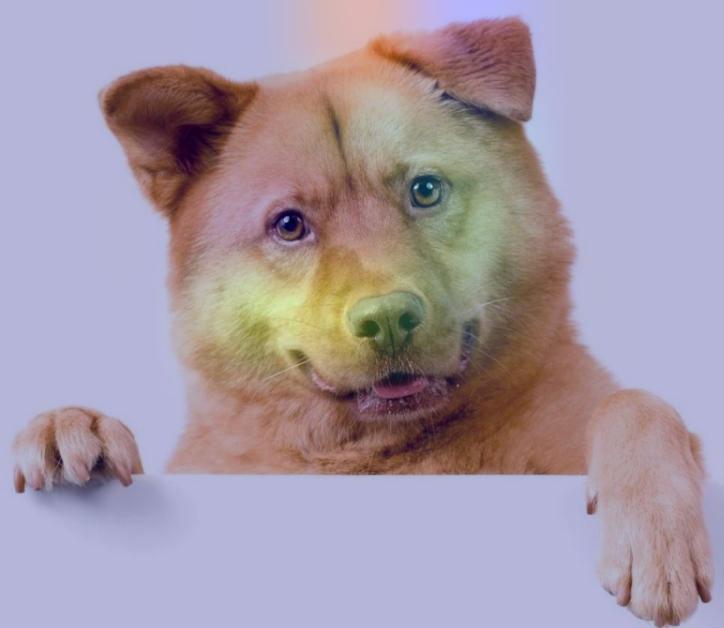}
    \caption{88\,\% Chow}
  \end{subfigure}
  \\*[1ex]
  \begin{subfigure}{0.96\linewidth}
    \centering
    \includegraphics[width=0.85\linewidth, viewport=46 26 530 650, clip]{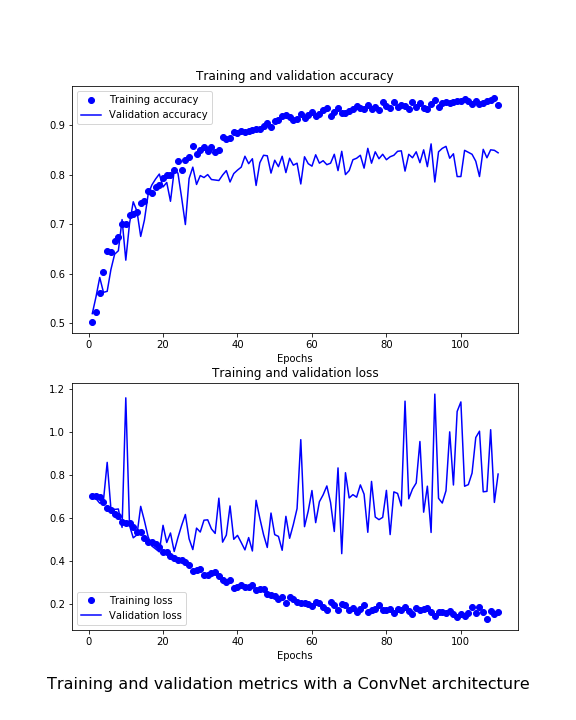}
    \caption{}
  \end{subfigure}
  \caption{\label{fig:inc-chow88}Deterministic ConvNet on HW-1: Grad-CAM explanation of the \emph{right} identification of the ``Chow'' (a), model's accuracy and validation loss curves (b).}
\end{figure}

\begin{figure}[htbp]\setlength{\tabcolsep}{0.7mm}
  \centering
  \begin{subfigure}{0.96\linewidth}
    \centering
    \includegraphics[width=0.7\linewidth]{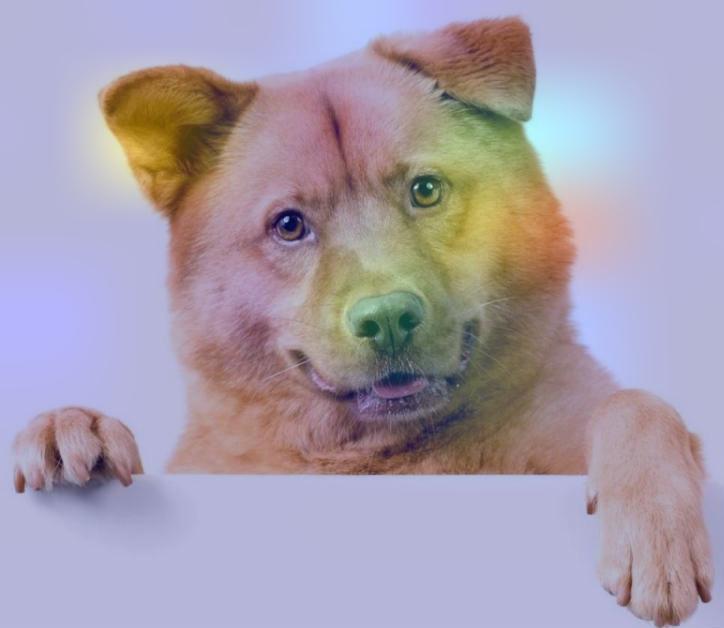}
    \caption{76\,\% Cat}
  \end{subfigure}
  \\*[1ex]
  \begin{subfigure}{0.96\linewidth}
    \centering
    \includegraphics[width=0.85\linewidth, viewport=46 26 530 650, clip]{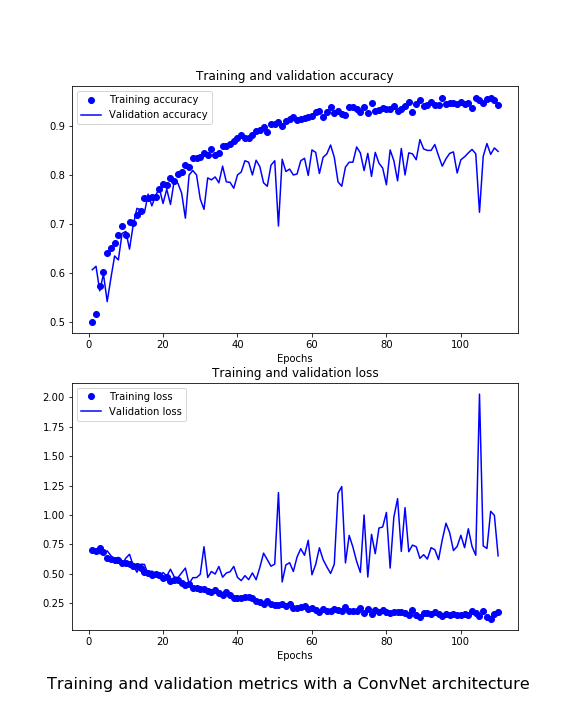}
    \caption{}
  \end{subfigure}
  \caption{\label{fig:inc-cat76}Deterministic ConvNet on HW-2: Grad-CAM explanation of the \emph{false} classification of the ``Chow'' as a ``Cat'' (a), model's accuracy and validation loss curves (b).}
\end{figure}

\begin{itemize}\setlength{\itemsep}{-0.4ex}\small
\item Number of layers: 12
\item Trainable parameters: 991,041
\item Training times:
  \begin{itemize}
  \item HW-1: ca. 2\,s / epoch 
  \item HW-3: ca. 3\,s / epoch 
  \item HW-4: ca. 20\,s / epoch 
  \item HW-5: ca. 22\,s / epoch 
  \end{itemize}
\end{itemize}
The deterministic ConvNet was developed as a proof of concept and it is
by no means an optimized
model. Therefore, the reproducibility of its Grad-CAM explanations but not their quality, are tested here. Classification explanations calculated here for a test image showing a Chow-Chow dog (image taken from~\cite{CholletGradCAMcav2021}) are depicted for two
different execution environments in~\cref{fig:inc-chow88} and~\cref{fig:inc-cat76}, respectively. The visual explanations remain stable (are reproducible) within each execution environment but they are, quite like the classification results, environment dependent.
The quality of the classifications is not the object of investigation for these simple, first deterministic models.
%

\subsection{Deterministic mini Xception}
A \emph{small} version of Xception,
the \emph{mini Xception-like} of Chollet~\cite{CholletDLwithPython2nd-2021}, has been chosen to be modified, so as to make its training deterministic. The
system configurations described in the previous section (\cref{l:EnvVars,l:seeding}), and the seed settings for the data augmentation had to be
applied also in this case. A couple of additional steps are however also necessary. The original model definition
is as given in~\cite{CholletDLwithPython2nd-2021}. The final version of the code, made deterministic for CPU environments, is displayed in~\cref{l:minidet}.
%
\begin{lstcode}[float=*ht,caption={\label{l:minidet}mini Xception: Deterministic (on CPU) training code.}]
    init_glorot_u = tf.keras.initializers.glorot_uniform

    inputs = keras.Input(shape=(PDIM, PDIM, 3))
    x = data_augmentation(inputs)
    x = keras.layers.experimental.preprocessing.Rescaling(1./255)(x)
    x = keras.layers.Conv2D(filters=32, kernel_size=5, use_bias=False,
                            kernel_initializer=init_glorot_u(seed=1))(x)

    for size in [32, 64, 128, 256, 512]:
        residual = x
        x = keras.layers.BatchNormalization()(x)
        x = keras.layers.Activation("relu")(x)
        x = keras.layers.SeparableConv2D(size, 3, padding="same", use_bias=False,
                                         depthwise_initializer=init_glorot_u(seed=1),
                                         pointwise_initializer=init_glorot_u(seed=2))(x)
        x = keras.layers.BatchNormalization()(x)
        x = keras.layers.Activation("relu")(x)
        x = keras.layers.SeparableConv2D(size, 3, padding="same", use_bias=False,
                                         depthwise_initializer=init_glorot_u(seed=1),
                                         pointwise_initializer=init_glorot_u(seed=2))(x)
        x = keras.layers.MaxPooling2D(3, strides=2, padding="same")(x)
        residual = keras.layers.Conv2D(size, 1, strides=2, padding="same", use_bias=False,
                                       kernel_initializer=init_glorot_u(seed=1))(residual)
        x = keras.layers.add([x, residual])

    x = keras.layers.GlobalAveragePooling2D()(x)
    x = keras.layers.Dropout(0.5, seed=7001)(x)
    outputs = keras.layers.Dense(1, activation="sigmoid")(x)
    model = keras.Model(inputs=inputs, outputs=outputs)
    model.compile(loss="binary_crossentropy", optimizer="rmsprop", metrics=["accuracy"])
\end{lstcode}
%
The model counts 718,849
trainable parameters (while the corresponding number given in~\cite{CholletDLwithPython2nd-2021} is actually the number of total parameters).
To enhance the layer function's determinism,
the parameters \lsti|depthwise_initializer| and the
\lsti|pointwise_initializer| of the \lsti|SeparableConv2D| are set
explicitly to the \lsti|glorot_uniform| initializer with a fixed
seed. The same change is applied to the
\lsti|kernel_initializer| in the function \lsti|Conv2D|.
This code is reproducible when executed in CPU environments without graphical support. Runtimes of the deterministic code for HW-2 are given in Table~\ref{tb:minixcp-cpu}.
\begin{table}[htbp]
  \caption{\label{tb:minixcp-cpu}Deterministic mini Xception on HW-2}
  \begin{minipage}[t]{\linewidth}
  \centering
  \vspace{-1mm}
  \begin{tabular}{| l |c |}
  \hline
  Single-thread & Multi-thread\\
  \hline
  ca. 330 s / Epoch & ca. 52 s / Epoch\\
 \hline
  \end{tabular}
  \vspace{-2mm}
  \end{minipage}%
\end{table}
Switching off parallelism of \lsti|Tensorflow| can be achieved by using the instructions given in~\cref{l:ptf}.
\begin{lstcode}[float=*ht,numbers=left,caption={\label{l:ptf}Switching off parallelism of Tensorflow.}]
tf.config.threading.set_inter_op_parallelism_threads(1)
tf.config.threading.set_intra_op_parallelism_threads(1)
\end{lstcode}

However, these instructions to turn off multi-threading had no practical
effect, when applied in the GPU environment. It was indeed not possible to make the GPU work in single-thread mode. To make mini Xception deterministic for GPU environments, two changes are necessary in the code listed in~\cref{l:minidet} which actually change the network:
%
\begin{enumerate}\setlength{\itemsep}{-0.3ex}
\item replace \lsti|SeparableConv2D| by \lsti|Conv2D|,
\item remove the line containing \lsti|keras.layers.add()|.
\end{enumerate}
%
The in this way resulting deterministic model however, has little in common with the original mini Xception.
%

\subsubsection*{Mini Xception Summary}
\begin{figure}[htbp]\setlength{\tabcolsep}{0.7mm}
  \centering
  \begin{subfigure}{0.96\linewidth}
    \centering
    \includegraphics[width=0.7\linewidth]{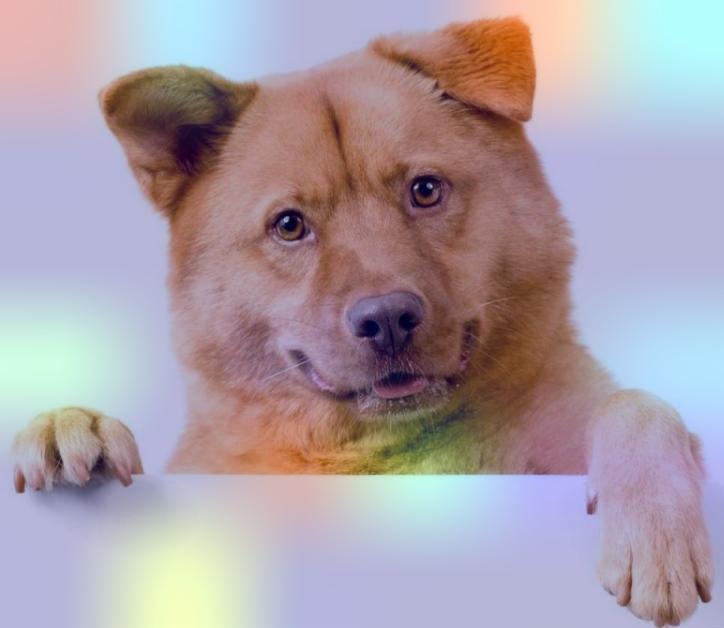}
    \caption{99\,\% Chow}
  \end{subfigure}
  \\*[1ex]
  \begin{subfigure}{0.96\linewidth}
    \centering
    \includegraphics[width=0.85\linewidth, viewport=46 26 530 650, clip]{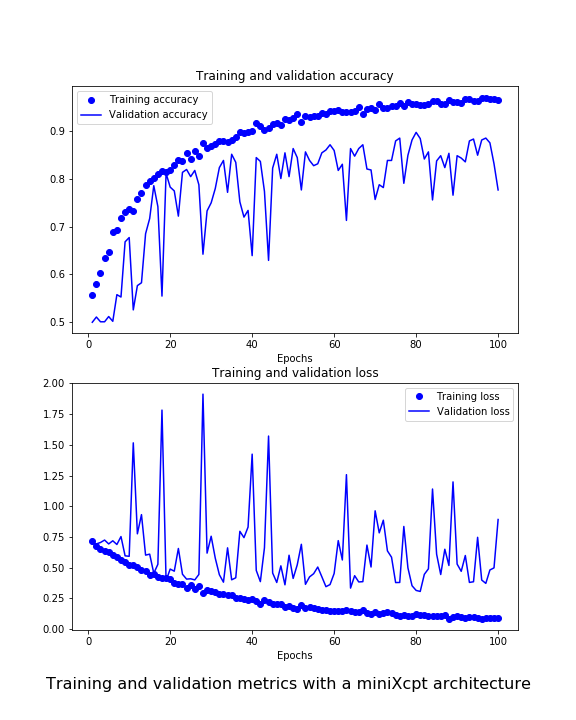}
    \caption{}
  \end{subfigure}
  \caption{\label{fig:incpt-chow99}Deterministic Mini Xception on HW-2: Grad-CAM explanation of the \emph{correct} classification as ``Chow'' (a), model's accuracy and validation loss curves (b).}
\end{figure}

\begin{figure}[htbp]\setlength{\tabcolsep}{0.7mm}
  \centering
  \begin{subfigure}{0.96\linewidth}
    \centering
    \includegraphics[width=0.7\linewidth]{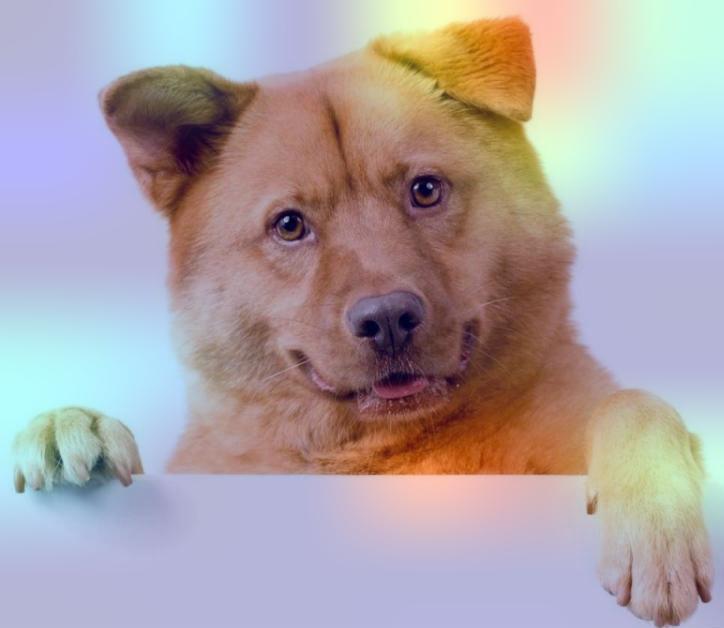}
    \caption{100\,\% Cat}
  \end{subfigure}
  \\*[1ex]
  \begin{subfigure}{0.96\linewidth}
    \centering
    \includegraphics[width=0.85\linewidth, viewport=46 26 530 650, clip]{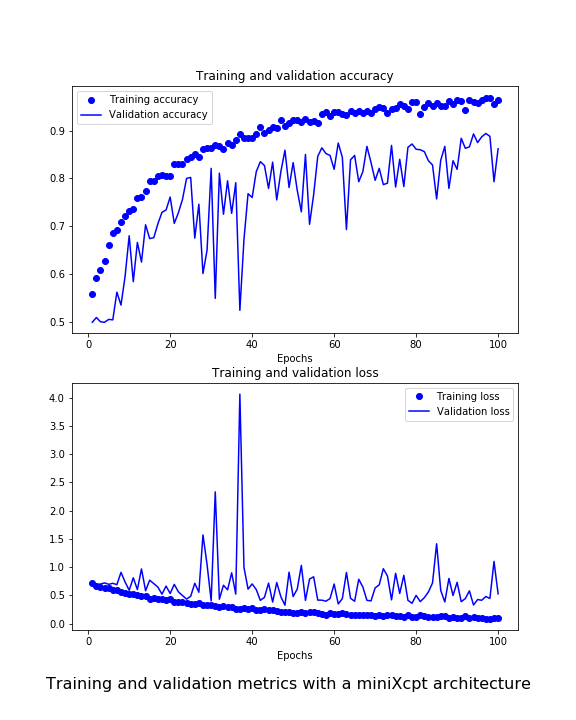}
    \caption{}
  \end{subfigure}
  \caption{\label{fig:incpt-cat100}Deterministic Mini Xception on HW-5: Grad-CAM explanation of the \emph{false} object classification as ``Cat'' (a), model's accuracy and validation loss curves (b).}
\end{figure}

\begin{itemize}\setlength{\itemsep}{-0.3ex}
\item Number of layers: 44
\item Trainable parameters: 718,849
\item Training times:
  \begin{itemize}
  \item HW-2: ca. 52\,s / epoch\\(to compare: single thread 330\,s / epoch)
  \item HW-4: ca. 54\,s / epoch
  \item HW-5: ca. 56\,s / epoch
  \end{itemize}
\end{itemize}
For a comparison: on HW-1, the non-deterministic training takes ca.~18\,s per epoch.
The deterministic mini Xception was developed as a proof of concept. Tests of the Grad-CAM explainability of the network's classifications have also been performed. In~\cref{fig:incpt-chow99} and ~\cref{fig:incpt-cat100} there are depicted the visual explanations of the model, as well as the training and validation loss curves respectively, after 100 epochs of training in two different execution environments.
The visual explanations remain stable (are reproducible) within each execution environment but they are, quite like the classification results, environment dependent.
The Grad-CAM process code used in this work is as given in Section 9.4.3 of~\cite{CholletDLwithPython2nd-2021}.

\section{Conclusions and future work}
\label{sec:Concl}
In this work, the non-reproducibility of the training of DL models in association with the employment of different HW environments (with and without GPU) for the execution
of the source code, has been discussed. The non-reproducibility of models, necessary implies the non-reproducibility of the explanations of their results, the quality of which is of vital importance, given the increasing dependency on ML for decision making in many sectors of life, like automotive, telecommunications, healthcare etc. Aside from the nondeterminism of training algorithms, core libraries like TensorFlow, CNTK, Theano, Pytorch and low-level libraries like for example cuDNN, are known to exhibit a so-called
\emph{implementation-level} variance across all their evaluated versions~\cite{varianceProb20}.
That means that even if the variance introduced by algorithmic factors like shuffled batch ordering, dropout regularization, data augmentation etc. becomes eliminated by using fixed random seeds, the core implementation variance will still be added to the results.
The development of deterministic code for two different CNN model architectures has been successfully achieved. This has been confirmed with test runs under the condition that the code is executed on similar CPU environments with no use of
graphic card support (here: AMD Zen architecture with identical number of cores). Additionally, the first of the models, the deterministic ConvNet, runs deterministically also with activated graphic card support.
The technical steps towards the creation of these results are described in detail. The code used and instructions are listed.
Using the Grad-CAM method, HW-dependent reproducible model classifications and their according reproducible explanations have been visually demonstrated for both models.
The two prototype networks have not been optimized in any way.
The number of training epochs, the training and validation metrics are arbitrary.
If they should be further optimized, each optimization step has to be rendered reproducible too.
Earlier performed tests had confirmed that the ready-trained Inception V3 and Xception models deliver HW-independent and reproducible results, whereby the results of the two models are different. It is well-known that DL models undergo further optimization stages after the end of their training and before they reach their final form. For example, ensemble techniques are used to teach a model how to average or combine predictions of multiple models. These steps are mostly intransparent and hardly reproducible too~\cite{LoganDeepConvNNEns2021,BrownleeEnsembleLearnNN2019,BorisEnsemblesML,MallReporCrisis2019}.
Reproducibility is essential also in the context of fault tolerance, iterative refinement,
debugging and optimization of adaptable models, especially for large scale and distributed workflow applications, like cloud computing platforms and Industry 4.0~\cite{NguyenMLDLsurvey2019}. The need for specific fault tolerance features for the properties of DL algorithms and their implementations has been already discussed elsewhere~\cite{faultToler,faultToler1,faultToler2}. Reproducibility is also the basis for developing comparison criteria and metrics for the objective evaluation of model properties, like robustness and trustworthiness~\cite{varianceProb20,HeilNature}. Visual model explanations are helpful for humans to create insights from data information but they are seldom unique, they are often non-intuitive or not consistent, as regards the selected features for the explanation. Moreover, DL models and their explanations are vulnerable to errors, for instance when slightest perturbations are introduced to the input image~\cite{Goodfellow2015ExplainHarnessAdvers}.
In addition, visual explanations don't make transparent how networks produce predictions, why gradients converge etc.~\cite{expmlfiddler}.
The reproducibility of ML training is expected to boost developments in XAI, with the scope to finally create humanly comprehensible, causal interpretations of decisions produced by complex DL models.
There exist no globally accepted measures of accuracy or correctness
for ML models, hence there is a need for a mathematical formalism to describe
properties, which models should prove to have and how to evaluate them. Approaches to an \emph{axiomatic formulation} of attribute-based predictions of DL networks, propose model properties with respect to the input-data features that must be fulfilled.  As such,
there have been named: completeness, sensitivity,
linearity, symmetry, continuity and implementation invariance~\cite{talyAxiomatic}.
For the future, it is intended to integrate some of these properties as boundary conditions in model implementations, which will have to be satisfied throughout the training process. An investigation,
if additional restrictions to the model can positively contribute to the model's reproducibility seems worth trying.
Implementation invariance is satisfied when two functionally equivalent networks find identical attributions for the same input image and baseline image~\cite{impinv}.
This work shows that a single implementation can be functionally non-equivalent to itself, if executed in different HW environments.
Future attempts to alleviate this condition should not completely disregard the investigation of also open architectures and standards for graphic cards.
%
{\small
\bibliographystyle{abbrv}
\bibliography{main}
}
\end{document}